\documentclass[letterpaper]{article} 
\usepackage[]{aaai2027}  
\usepackage[hyphens]{url}  
\usepackage{graphicx} 
\usepackage{natbib}  
\usepackage{caption} 
\usepackage{algorithm}
\usepackage{algorithmic}

\usepackage{newfloat}
\usepackage{listings}
\DeclareCaptionStyle{ruled}{labelfont=normalfont,labelsep=colon,strut=off} 
\floatstyle{ruled}
\newfloat{listing}{tb}{lst}{}
\floatname{listing}{Listing}

\usepackage{booktabs}
\usepackage{multirow}

\usepackage{amsmath}
\usepackage{amssymb}
\usepackage{tikz}
\usepackage{soul}
\usepackage{xspace}
\usepackage{tcolorbox}
\usepackage{alltt}
\usepackage{makecell}

\title{Mixture-of-Expert Blocks Contain Strong Hallucination Detection Signals}
\author{
    Joao Fonseca\textsuperscript{\rm 1},
    Rodrigo Rodrigues\textsuperscript{\rm 1, \rm 2},
    Paolo Romano\textsuperscript{\rm 1, \rm 2}\corresponding
}
\affiliations{
    \textsuperscript{\rm 1}INESC-ID, Rua Alves Redol, 9, Lisbon, 1000-029 Portugal \\
    \textsuperscript{\rm 2}Instituto Superior Técnico\\
    \{joaofonseca, rodrigo.rodrigues, romano\}@inesc-id.pt
}

\newcommand*{\method}{\texttt{InnerExpert}\xspace}
\newcommand{\ie}{\textit{i.e., \xspace}}
\newcommand{\eg}{\textit{e.g., \xspace}}

\newcommand{\stepcircled}[1]{%
  \tikz[baseline=(char.base)]{
    \node[shape=circle,draw=black,fill=black,inner sep=1pt] (char) {\textcolor{white}{\textbf{#1}}};
  }%
}

\begin{document}

\maketitle

\begin{abstract}
  Despite their widespread use, Large Language Models (LLMs) remain limited by
  a fundamental problem: the generation of plausible but false content, known
  as \emph{hallucinations}. Most existing detection methods operate at the
  answer or sentence level, yet per-token detection is essential for localizing
  hallucinated spans and enabling fine-grained interventions. In this paper,
  we explore the use of the Mixture-of-Experts (MoE) paradigm to address this
  gap. In MoE architectures, a single forward pass activates a sparse subset of
  experts (\ie distinct feedforward networks per layer) via a routing
  mechanism, producing internal signals (\eg router entropy, expert
  disagreement, and expert usage patterns) that are unavailable in dense
  architectures and have not been previously exploited for hallucination
  detection. To this end, we introduce \method, the first method to leverage
  these MoE-specific signals for per-token hallucination detection. \method
  combines routing-level and standard transformer signals into compact
  per-token feature vectors, classified by a lightweight detector trained on
  labels produced by an LLM-as-a-judge pipeline, which enables continuous model
  updates without manual annotation. Our results show that \method
  outperforms existing methods across five datasets and two MoE architectures,
  achieving up to 0.91 answer-level and 0.76 token-level AUROC, while
  requiring only a single forward pass.
\end{abstract}

\begin{links}
    \link{Code}{https://github.com/joaopfonseca/InnerExpert-Hallucination-Detection}
\end{links}

\section{Introduction}\label{sec:introduction}

\begin{figure}[t!]
  \centering
  \includegraphics[width=\linewidth]{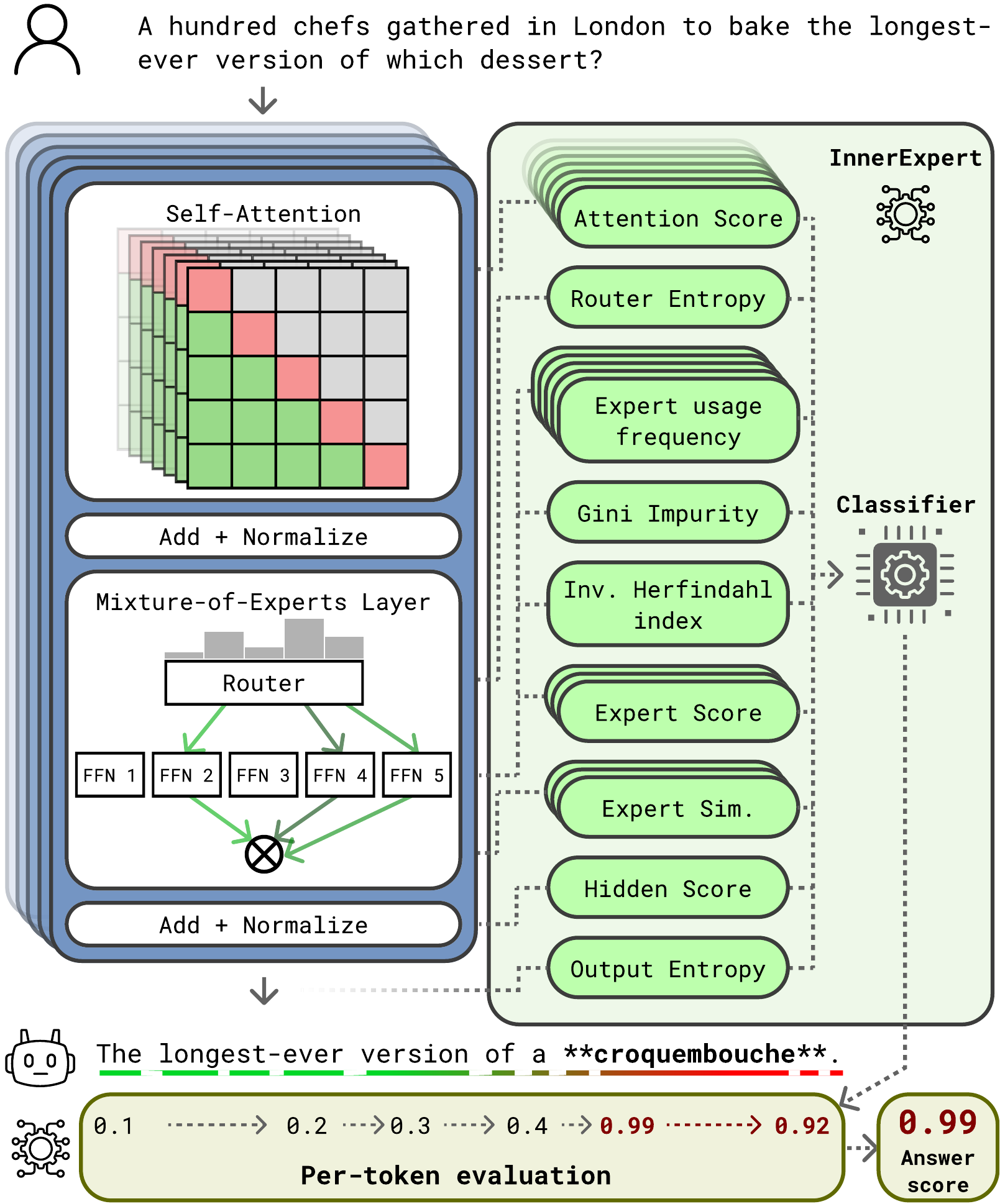}
  \caption{
    Overview of \method. A single forward pass through the MoE model exposes
    internal signals (\ie router distributions, per-expert hidden states, and
    expert usage patterns) which are combined into per-token hallucination
    scores.
  }\label{fig:overview}
\end{figure}

The reliability of Large Language Models (LLMs) is often questioned due to the
risk of generating plausible but false content, commonly referred to as
\emph{hallucination}~\cite{alansari2026large}. Hallucinations are particularly
prevalent when LLMs are prompted for information that is outside their training
distribution (\eg post-training events or highly specialized
knowledge)~\cite{vu2024freshllms}. Despite the importance of this problem, detecting
hallucinations reliably and efficiently remains an open challenge. 

The Mixture-of-Experts (MoE) paradigm has become prevalent in frontier
open-weight LLMs, as it allows for larger parameter counts and greater model
performance at a feasible inference cost~\cite{cai2025survey}. A distinctive property of MoE models
(as opposed to dense models) is that a forward pass selects, via a routing
function, a sparse subset of \emph{experts} (\ie feedforward networks) per
layer~\cite{fedus2022switch}. This process produces a large amount of signals,
most of which are unavailable in dense architectures, such as router entropy,
disagreement between experts' hidden states, and the distribution of expert
usage over a sequence.
Figure~\ref{fig:overview} illustrates this approach at a high level.

Although this
approach may allow the development of cheap and efficient hallucination
detection methods, to the best of our knowledge, no prior work leverages
MoE-specific internal signals to detect hallucinations in LLMs. These signals
have been shown to relate to epistemic
uncertainty~\cite{pavlitska2025extracting}, providing a theoretical motivation
for their use in hallucination detection; we develop this connection in
Appendix~\ref{sec:epistemic-uncertainty}. Moreover, because MoE routing
decisions are made at each token, these signals are naturally aligned with
per-token granularity, enabling fine-grained localization of hallucinated
spans. However, most existing methods operate at the answer
or sentence level and do not provide such capability.

Existing hallucination detection methods can be grouped into three paradigms.
Sampling-based approaches, such as SelfCheckGPT~\cite{manakul2023selfcheckgpt}
and Semantic Uncertainty~\cite{kuhn2023semantic}, probe consistency across
samples (obtained via multiple forward-passes), which is expensive at inference time. INSIDE~\cite{chen2024inside} is a sampling-based method that also utilizes hidden states over multiple responses.
Internal-signal methods, such as LLM-Check~\cite{sriramanan2024llm}, and MIND~\cite{su2024unsupervised}, operate on
hidden states and attention in a single pass, but generally employ a limited
subset of these signals and ignore the MoE-specific routing structure.
Trainable detectors, such as HaluNet~\cite{tong2025halunet} and
FacLens~\cite{wang2025faclens}, learn classifiers using dense transformer
signals over entire answers, but none of these operate at the token level. A
complementary line of work on uncertainty estimation decomposes predictive
uncertainty into aleatoric and epistemic
components~\cite{hullermeier2021aleatoric, yadkori2024believe}, providing
theoretical motivation for internal-signal approaches, while
\citet{pavlitska2025extracting} show that MoE-specific signals capture
epistemic uncertainty in the vision domain. Additional details on the above approaches are provided in Appendix~\ref{sec:related-work}. \method
addresses these gaps by systematically extracting and combining routing-level
signals with standard transformer signals into a unified hallucination
detection framework.

\method combines the internal-signal and trainable-detector paradigms. It
preserves the single-pass efficiency of internal-signal methods while adopting
the trainable-classifier approach of detectors, leveraging the fact that MoE models
expose additional routing-level signals, such as router entropy, expert
disagreement, and expert usage patterns. To the best of our knowledge, no prior
method explores these signals for hallucination detection in text generation.
Furthermore, most existing methods operate at the answer or
sentence level; per-token hallucination detection, which enables fine-grained
localization of hallucinated spans, remains largely underexplored. 
\textbf{\method combines MoE-specific signals at
per-token granularity, addressing both gaps simultaneously.} 
Importantly, we show that \method achieves state-of-the-art 
performance in
hallucination detection using simple classifiers, \ie without employing complex
architectures such as in HaluNet or FacLens, which we leave to future work. 
In summary, our contributions are as follows:

\noindent
\stepcircled{1} We propose \method, a single-pass, per-token hallucination
detector that observes the model's internal states and leverages MoE-specific internal signals, requiring no modifications to the host LLM or additional sampling.

\noindent
\stepcircled{2} We formalize an unsupervised training approach for per-token
hallucination detection, designed to train \method, based on LLM-based
labeling of generated answers against reference evidence.

\noindent
\stepcircled{3} We demonstrate that \method achieves an average competitive performance of up to 0.91 AUROC at the answer-level and 0.76 AUROC at the token-level. 
in the hallucination detection task across five datasets and two different
open-weight MoE architectures, against baselines spanning the
sampling-based, internal-signal, and trainable paradigms.

\noindent
\stepcircled{4} We formalize an inventory of six MoE-specific
signals alongside the standard hidden state and attention scores used in prior work,
enabling a systematic study of which routing signals carry
hallucination-discriminative information.

\section{Background}\label{sec:background}

\paragraph{Hallucination in LLMs.}
We consider an autoregressive LLM with parameters $\theta$ that, given a prompt (\ie sequence of input tokens)
$\mathbf{x}=(x_1,\dots,x_U)$, generates a sequence $\mathbf{y}=(y_1,\dots,y_T)$ by sampling
each token from $p_\theta(y_t\mid \mathbf{x}, y_{<t})$. A \emph{hallucination}
corresponds to a subset of fluent and internally coherent tokens
$\mathbf{y}_h\subseteq\mathbf{y}$ that are factually incorrect or
unsupported~\cite{alansari2026large}.\footnote{Some incorrect outputs reflect
the \emph{faithful propagation} of erroneous content already present in the
training data or in retrieved context, rather than a generation failure per se;
both sampling-based and internal-signal detectors treat such cases as
low-uncertainty by design. The temporal-out-of-distribution evaluation we
employ (Section~\ref{sec:experiments}) mitigates this conflation, since
incorrect answers to post-training questions are predominantly fabrications.}

\paragraph{Mixture-of-Experts and Its Internal Signals.}
A MoE replaces each feedforward sublayer of a standard transformer architecture
with a sparse mixture of $N$ expert networks $\{E_1,\dots,E_N\}$ and a gating
(router) network, parameterized by $\mathbf{W}_l^g$, that selects a 
subset of $k$ experts per layer, per token~\cite{fedus2022switch}. Let
$\mathbf{h}_{t,l}\in\mathbb{R}^d$ denote the hidden state at
token position $t$ entering layer $l$. Within layer $l$, the self-attention
sublayer and subsequent post-attention layernorm produce an intermediate
representation $\mathbf{h}'_{t,l}\in\mathbb{R}^d$ that serves as input to the
MoE sublayer. The router produces a probability distribution over experts from this
representation,
\begin{equation}\label{eq:router}
g_l(\mathbf{h}'_{t,l}) = \mathrm{softmax}\!\left(\mathbf{W}_l^g\, \mathbf{h}'_{t,l}\right) \in \mathbb{R}^N,
\end{equation}
a subset $\mathcal{S}_{t,l}\subseteq\{1,\dots,N\}$ of $k$ experts is then
selected (typically by top-$k$), and the MoE sublayer contribution is the
weighted combination of expert outputs, added to the residual stream to form
the hidden state entering the next layer:
\begin{equation}\label{eq:moe-output}
\begin{split}
\mathbf{h}_{t,l+1} &= \mathbf{h}_{t,l} + \mathrm{MoE}_l(\mathbf{h}'_{t,l}), \\
\mathrm{MoE}_l(\mathbf{h}'_{t,l}) &= \sum_{i \in \mathcal{S}_{t,l}} g_{l,i}(\mathbf{h}'_{t,l})\, E_i(\mathbf{h}'_{t,l}).
\end{split}
\end{equation}
A single forward pass of an MoE model therefore exposes internal signals that
are unavailable in dense architectures. 

\paragraph{Hallucination detection.}
\method aims to produce a score $s(y_t)\in [0, 1]$ for each generated token
$y_t$ and, by aggregation, an answer-level score
$S(\mathbf{y})=\mathrm{agg}(\{s(y_t)\}_{t=1}^{T})$ that indicates the
likelihood that $y_t$ (or $\mathbf{y}$) is hallucinated. We compute $s(y_t)$ from the internal signals
$\mathcal{I}(y_t)$ that the model exposes during a single forward pass:
aggregate and per-expert hidden states, attention weights, output logits,
cumulative routing distributions and expert activations. A token is flagged as hallucinated when $s(y_t)$ exceeds a threshold $\tau$, which can be tuned to
trade off precision for recall using standard methods; the answer-level
prediction is obtained by aggregating per-token scores (\eg by mean or max) and
thresholding. The specific signals and scoring function used in \method are
defined in Section~\ref{sec:method}.

\section{\method}\label{sec:method}

\method combines
MoE-specific internal signals and
combines them via a meta-expert (\ie a trained classifier) into per-token hallucination scores (as summarized in
Figure~\ref{fig:overview}). Since the host model's signals are very
high-dimensional, the Expert is intentionally fed a compact set of scores
derived from the internal signals, rather than the raw hidden states, attention
matrices and router logits. Although it is not required to modify the host model and no additional generations are needed beyond the forward
passes that produce the answer under evaluation, some practical implementation
challenges must be considered to achieve the full observability of the host model, which we detail in
Appendix~\ref{sec:implementation-details}.

\subsection{Architecture}\label{subsec:architecture}

During generation, \method intercepts the model's forward pass at each decoding
step and collects a comprehensive set of internal signals at every layer.
We distinguish between \emph{standard signals}, available in any
transformer-based architecture, and \emph{MoE-specific signals}, which arise
from the routing structure and are absent in dense architectures. All signals
are computed cumulatively over the generated prefix $y_{\leq t}$, yielding a
per-token value at each generation step~$t$.

\paragraph{Standard signals.}
We adapt two complementary signals from LLM-Check~\cite{sriramanan2024llm},
which proposes single-pass internal signals that correlate with
hallucination in prior work. The hidden state score captures
representational redundancy across the generated sequence, while the
attention score captures the model's focus patterns; both are computed
cumulatively over the prefix, aligning naturally with per-token
granularity:

\emph{Hidden state score.}
For each layer $l$, let $\mathbf{H}_{t,l} \in \mathbb{R}^{t \times d}$ denote
the matrix of hidden states $\{\mathbf{h}_{1,l}, \dots, \mathbf{h}_{t,l}\}$
collected up to position $t$. Following
\citet{sriramanan2024llm}, we compute the centered covariance
$\boldsymbol{\Sigma}_{t,l} = \mathbf{H}_{t,l}^\top \mathbf{J}\, \mathbf{H}_{t,l}
+ \alpha \mathbf{I}$, where $\mathbf{J} = \mathbf{I} - \frac{1}{d}\mathbf{1}\mathbf{1}^\top$
centering over the hidden dimension and $\alpha > 0$ is a regularization
constant (we use $\alpha = 0.001$). The hidden state score is the mean log-singular-value of
$\boldsymbol{\Sigma}_{t,l}$:
\begin{equation}\label{eq:hidden-score}
\phi^{\mathrm{hid}}_{t,l} = \frac{1}{t}\sum_{j=1}^{t} \log\,\sigma_j\!\left(\boldsymbol{\Sigma}_{t,l}\right),
\end{equation}
where $\sigma_j(\cdot)$ denotes the $j$-th singular value. Low values indicate
low-rank (redundant) hidden representations, associated with confident
generation~\cite{sriramanan2024llm}.

\emph{Attention score.}
Let $\mathbf{A}_{t,l,h} \in \mathbb{R}^{t \times t}$ be the attention matrix of
head $h$ at layer $l$ up to position $t$. The attention score is the
log-cumulative sum of the diagonal entries of the attention kernel:
\begin{equation}\label{eq:attention-score}
\phi^{\mathrm{att}}_{t,l,h} = \sum_{j=1}^{t} \log\!\left(\mathbf{A}_{t,l,h}[j,j] + \epsilon\right),
\end{equation}
where $\epsilon$ is a small constant for numerical stability. The diagonal of
the attention kernel equals its eigenvalues, so this quantity approximates the
log-determinant of the attention similarity matrix~\cite{sriramanan2024llm}.

\paragraph{MoE-specific signals.}
The routing structure of MoE models exposes additional signals unavailable in
dense architectures. We define six signals covering three complementary
aspects of routing behavior: routing uncertainty (router entropy),
expert-level representation quality (expert hidden score, expert
similarity), and routing distribution dynamics (expert usage
distribution, Gini impurity, inverse Herfindahl). This selection is
grounded in the connection between MoE routing and epistemic
uncertainty~\cite{pavlitska2025extracting}, detailed in
Appendix~\ref{sec:epistemic-uncertainty}:

\emph{Router entropy.}
The entropy of the gating distribution at each layer,
\begin{equation}\label{eq:router-entropy}
\phi^{\mathrm{rout}}_{t,l} = \mathrm{H}\!\left[g_l(\mathbf{h}'_{t,l})\right]
= -\sum_{i=1}^{N} g_{l,i}(\mathbf{h}'_{t,l})\,
\log g_{l,i}(\mathbf{h}'_{t,l}),
\end{equation}
is high when the router is uncertain about which expert should handle the
token, suggesting the input falls in a region where the model's knowledge may be
insufficient.

\emph{Expert hidden score.}
Each selected expert $i \in \mathcal{S}_{t,l}$ produces an output
$\mathbf{e}_{t,l,i} = E_i(\mathbf{h}'_{t,l})$. We compute the hidden state
score (Eq.~\ref{eq:hidden-score}) independently for each expert's output
sequence and aggregate via the routing weights:
\begin{equation}\label{eq:expert-hidden-score}
\phi^{\mathrm{exp\text{-}hid}}_{t,l}
= \sum_{i \in \mathcal{S}_{t,l}} g_{l,i}(\mathbf{h}'_{t,l})\,
\phi^{\mathrm{hid}}_{t,l,i},
\end{equation}
where $\phi^{\mathrm{hid}}_{t,l,i}$ is the hidden state score computed from the
per-expert output sequence $\{\mathbf{e}_{1,l,i}, \dots, \mathbf{e}_{t,l,i}\}$.
This measures whether the activated experts individually produce confident
representations.

\emph{Expert similarity.}
Low pairwise similarity among the activated experts' outputs indicates
disagreement, analogous to ensemble disagreement, which has been shown to
capture epistemic uncertainty~\cite{pavlitska2025extracting}. We compute the
routing-weighted average of pairwise cosine similarities:
\begin{equation}\label{eq:expert-similarity}
\phi^{\mathrm{sim}}_{t,l}
= \sum_{i \in \mathcal{S}_{t,l}} \sum_{j \in \mathcal{S}_{t,l}}
g_{l,i}(\mathbf{h}'_{t,l})\, g_{l,j}(\mathbf{h}'_{t,l})\,
\cos\!\left(\mathbf{e}_{t,l,i},\, \mathbf{e}_{t,l,j}\right).
\end{equation}
with low values indicating that the activated experts produce inconsistent
representations for the current token.

\emph{Expert usage distribution.}
We maintain a cumulative count of expert selections up to position $t$,
weighted by routing probabilities and normalized into a distribution:
\begin{equation}\label{eq:expert-usage}
\mathbf{u}_{t,l} = \frac{\sum_{t'=1}^{t} \mathbf{c}_{t',l}}
{\left\|\sum_{t'=1}^{t} \mathbf{c}_{t',l}\right\|_1}
\in \Delta^{N-1},
\end{equation}
where
\begin{equation}\label{eq:expert-usage-weights}
c_{t',l,i} = \begin{cases}
    g_{l,i}(\mathbf{h}'_{t',l}) & \text{if } i \in \mathcal{S}_{t',l} \\
    0                           & \text{otherwise}
\end{cases}
\end{equation}
retains the routing probability for each selected expert and zeros out the
rest. This distribution captures how routing has evolved over the generated
sequence.

\emph{Gini impurity.}
The Gini impurity of the expert usage distribution,
\begin{equation}\label{eq:gini}
\phi^{\mathrm{gini}}_{t,l} = 1 - \sum_{i=1}^{N} u_{t,l,i}^2,
\end{equation}
is high when many experts are selected equally often (scattered routing) and
low when few experts dominate.

\emph{Inverse Herfindahl index.}
The inverse Herfindahl index, or effective number of experts,
\begin{equation}\label{eq:herfindahl}
\phi^{\mathrm{herf}}_{t,l} = \left(\sum_{i=1}^{N} u_{t,l,i}^2\right)^{-1},
\end{equation}
captures how many experts are effectively contributing to the generation.
A value close to $1$ indicates a single expert dominates; a value close to $N$
indicates uniform usage.

These signals expose the model's routing behavior and the consistency of the
experts' hidden states at the token level. A summary of the metrics computed
and their dimensionalities (for a single forward pass) is available in
Appendix~\ref{sec:method-details}. To the best of our knowledge, no prior work
leverages these signals for hallucination detection.

\paragraph{Feature Assembly.} In order for the signals to be used as input to
the hallucination expert, for each generated token $y_t$, the per-layer signals
are flattened into a single feature vector in a fixed order,
\begin{equation}\label{eq:feature-vector}
\begin{split}
\Phi_t = \{&\phi^{\mathrm{hid}}_{t,l}, \phi^{\mathrm{att}}_{t,l,1}, \dots,
\phi^{\mathrm{att}}_{t,l,h},
\phi^{\mathrm{rout}}_{t,l}, \\
  & \phi^{\mathrm{exp\text{-}hid}}_{t,l},
\phi^{\mathrm{sim}}_{t,l}, \phi^{\mathrm{gini}}_{t,l},
\phi^{\mathrm{herf}}_{t,l}, \mathbf{u}_{t,l}\}.
\end{split}
\end{equation}
Signals are collected only for generated tokens (not the prompt
prefix). During classifier training, all features except the expert usage
distribution (which is already bounded in $[0,1]$) are standardized via
z-score normalization; the expert usage ratios are passed through unscaled.

\paragraph{Per-token hallucination scoring.} As stated in
Section~\ref{sec:background}, our goal is to produce a scoring function $s(y_t,
\Phi_t) \in [0,1]$ that predicts the probability that a generated token $y_t$
is hallucinated. In \method, $s$ is a lightweight classifier trained on labeled
data (Section~\ref{subsec:training-supervision}). Any standard ML classifier
can be used; we evaluate five families (Logistic Regression, Random Forest,
XGBoost, a multilayer perceptron, and a Transformer encoder). Full
hyperparameter grids are provided in Appendix~\ref{sec:implementation-details}.

\paragraph{Answer-level aggregation.}
Finally, the answer-level score is obtained by aggregating per-token scores:
\begin{equation}\label{eq:answer-score}
S(\mathbf{y}) = \mathrm{agg}\!\left(\{s(\Phi_t)\}_{t=1}^{T}\right),
\end{equation}
where $\mathrm{agg}$ is the mean or max operator. In our implementation we use
the max operator, \ie we consider an answer to be hallucinated if any of its
tokens is predicted to be hallucinated. Therefore, an answer is flagged as
hallucinated when $S(\mathbf{y})$ exceeds a threshold $\tau$. We discuss
potential uses of $\tau$ in future work for domain certification and controlled
text generation in Appendix~\ref{sec:future-work}.

\subsection{Training Supervision}\label{subsec:training-supervision}

We aim to set up a training regime for \method that does not require manual
annotation of hallucinations and allows for continuous model updates without
manual effort. To do this, we leverage the RealTime QA
platform~\cite{kasai2023realtime} to obtain a stream of question-answer pairs
$(\mathbf{x}, \mathbf{y})$ with reference evidence $\mathbf{e}$, where
$\mathbf{y}$ is generated by the host model. In our experiments, we use
questions posed between January 1st 2024 and December 31st 2025 for training.
Afterwards, we form per-token
hallucination labels with LLM-as-a-judge evaluation~\cite{zheng2023judging}. In
our experiments, we use GLM-5.1 as the judge. For each question, the host model
generates two answers: one with access to the reference evidence, to emulate
retrieval-augmented generation, and one without, which is more likely to
contain hallucinations.

\paragraph{Answer-level labels.}
For each generated answer, an LLM judge compares the answer against reference
evidence and produces (i) a binary hallucination label and (ii) a list of
hallucinated span strings (exact substrings from the answer). The judge is
prompted to return structured JSON with a label field ($1 = $ hallucinated, $0
= $ grounded) and a field listing the hallucinated substrings.

\paragraph{Token-level labels.} 
The hallucinated spans are mapped to per-token binary labels: a token $y_t$ is
labeled as hallucinated ($\ell_t = 1$) if its character span overlaps with any
hallucinated span, and as grounded ($\ell_t = 0$) otherwise. This produces a
token-level supervision signal aligned with the feature vector $\Phi_t$.

\paragraph{Train/validation split.} The labeled data is split into training and
validation sets stratified by hallucination rate, grouped by question
identifier to prevent leakage between answers to the same question. The
validation set is used to compute model performance while performing
hyperparameter grid search on the classifier. The best-found hyperparameter set per model is used
to train a final model on the combined training and validation sets. The final
  model is evaluated on a held-out test set, and the threshold $\tau$ is selected
to maximize F1-Score over all the labeled data. Out-of-distribution evaluation
is performed on data from a temporally disjoint period (see
Section~\ref{sec:experiments}). We validate the quality of the LLM-as-judge
labels against human annotations in
Appendix~\ref{subsec:llm-judge-validation}.

\section{Experiments}\label{sec:experiments}

Our experimental procedure is organized around four research questions (RQs):
\textbf{RQ1:} Can \method serve as a reliable indicator of
per-token hallucination? \textbf{RQ2:} How does \method compare to existing
hallucination detection baselines spanning the sampling-based, internal-signal,
and trainable-detector paradigms? \textbf{RQ3:} Which signals contribute most
to hallucination detection performance? \textbf{RQ4:} How reliably does the
LLM-as-a-judge labeling pipeline produce ground-truth hallucination labels?
The LLM-as-a-judge evaluation is detailed in
Appendix~\ref{subsec:llm-judge-validation}.

\paragraph{Baselines.} We compare \method against baselines spanning the three
paradigms identified in Section~\ref{sec:introduction}. These baselines are
also briefly described in Appendix~\ref{sec:related-work}. For sampling-based
methods, we generate $K{=}5$ stochastic samples per question with temperature $0.7$ and top-$p{=}0.9$.

\noindent \emph{$\bullet$ Sampling-based.} We evaluate two variants of
SelfCheckGPT~\cite{manakul2023selfcheckgpt}: the NLI-based variant and a
prompt-based variant. We also evaluate Semantic
Uncertainty~\cite{kuhn2023semantic} and Semantic Energy~\cite{ma2025semantic}.

\noindent \emph{$\bullet$ Internal-signal.} We evaluate two scoring variants from
LLM-Check~\cite{sriramanan2024llm}, attention score and hidden score, evaluated
independently with per-token scores aggregated to answer level via mean and
max. As standard floor baselines, we include the logit entropy (\ie per-token
Shannon entropy of the model's output distribution) and perplexity.

\noindent \emph{$\bullet$ Trainable.} We evaluate HaluNet~\cite{tong2025halunet}, a multi-branch
neural detector that fuses token-level log-likelihoods, entropy scores, and
hidden-state embeddings. HaluNet is trained with binary cross-entropy loss. As
no open-source implementation of HaluNet is available, we provide a
re-implementation based on the architecture described in the paper; full
hyperparameters are provided in Appendix~\ref{sec:implementation-details}.

\paragraph{Datasets.} We train \method and all baselines exclusively on
RealtimeQA~\cite{kasai2023realtime} data from January 2024 to December 2025,
labeled via the unsupervised LLM-as-judge pipeline described in
Section~\ref{subsec:training-supervision}. We evaluate on two categories of
test data: 
\begin{enumerate}
  \item A temporally out-of-distribution subset of RealtimeQA from January to
  June 2026 (365 questions), which tests whether the detector generalizes to
  post-training events.
  \item Four additional QA datasets: SQuAD~\cite{rajpurkar2016squad},
  TruthfulQA~\cite{lin2022truthfulqa},
  NQ-Open~\cite{kwiatkowski2019natural,lee2019latent}, and
  FreshQA~\cite{vu2024freshllms}, sampled at 200 questions each, which test
  cross-dataset generalization to different question distributions. 
\end{enumerate}
All datasets contain English-language questions and answers. RealtimeQA and
SQuAD additionally provide reference evidence; TruthfulQA, NQ-Open, and FreshQA
do not. We leave multilingual hallucination detection to future work. Full
dataset descriptions and summary statistics are provided in
Appendix~\ref{sec:datasets-description}.

\paragraph{Host models.} We use two open-weight MoE models as host architectures:
OLMoE-1B-7B-0924-Instruct~\cite{muennighoff2025olmoe} (64 experts, 8 active per token) and
Gemma-4-26B-A4B-it~\cite{gemmateam2026gemma} (128 experts plus 1 shared, 8 active per token).
Although Gemma 4 is a multimodal model, we use it only for text-to-text
generation.

\paragraph{Generation settings.} For each question, the host model generates answers
using greedy decoding with a maximum of 65 new tokens. For datasets that
provide reference evidence (RealtimeQA and SQuAD), we generate two answers per
question: one without evidence (base) and one with retrieved evidence (RAG
simulation). For datasets without evidence (TruthfulQA, NQ-Open, and FreshQA),
we generate only the base answer. This procedure is applied identically during
both training and evaluation. For sampling-based baselines, we generate
stochastic samples using the settings described in ``Baselines''.

\paragraph{Hyperparameters.} The \method classifier is selected via grid search
over five different classifiers: Logistic Regression, Random Forest, XGBoost,
and a multilayer perceptron (MLP). Hyperparameters are scored by F1 on a
single validation fold (10\% of training data, grouped by question
identifier), and the best-performing hyperparameter sets per classifier is
refit on the combined training and validation data. The decision threshold
$\tau$ is then F1-optimized on this combined set via a precision-recall curve
sweep. The complete hyperparameter grids are provided in Appendix~\ref{sec:additional-experiment-details}.

\paragraph{Performance metrics.} We evaluate all methods at both answer
level and
token level. For threshold-independent comparison, we report AUROC.
For threshold-dependent comparison, in
Appendices~\ref{sec:per-token-cont},~\ref{sec:comparison-baselines-cont}
and~\ref{sec:signal-analysis-cont} we report F1-Scores. All
metrics are computed using the labels produced by the LLM-as-judge pipeline
described in Section~\ref{subsec:training-supervision}.

\paragraph{Hardware and software.} All experiments were conducted on a machine
with an AMD EPYC 9555P 64-core processor (128 threads), 258\,GiB RAM, and two
NVIDIA RTX PRO 6000 Blackwell Max-Q Workstation Edition GPUs (96\,GiB VRAM
each), using Ubuntu 26.04 LTS. The \method classifier and baselines were
trained using scikit-learn 1.8 and PyTorch 2.10. Hallucination labels were
generated using GLM-5.1 accessed via the DeepInfra API.

\subsection{Per-token hallucination detection (RQ1)}\label{subsec:per-token}

\begin{table}
  \caption{Token-level results (AUROC) across datasets and host models.}
\label{tab:token_level_results}
\centering
\footnotesize
\setlength{\tabcolsep}{3pt}
\begin{tabular}{@{}lccccccc@{}}
\toprule
  & FQA & NQO & RTQA & SQuAD & TQA & Rank & Avg \\
\midrule
  \multicolumn{8}{c}{Gemma-4-26B} \\
\midrule
 IE (LR)  & 0.705 & 0.635 & 0.759 & 0.785 & 0.740 & 3.0 & 0.725 \\
 IE (MLP)  & 0.710 & 0.662 & 0.756 & \textbf{0.806} & 0.715 & 2.6 & 0.730 \\
 IE (RF) & 0.638 & 0.651 & 0.749 & 0.788 & 0.724 & 3.6 & 0.710 \\
 IE (Transf.) & 0.656 & 0.619 & 0.716 & 0.760 & 0.722 & 4.6 & 0.694 \\
 IE (XGB) & \textbf{0.711} & \textbf{0.678} & \textbf{0.805} & 0.802 & \textbf{0.767} & \textbf{1.2} & \textbf{0.753} \\
 LC (att.) & 0.310 & 0.331 & 0.370 & 0.332 & 0.223 & 8.0 & 0.313 \\
 LC (hid.) & 0.525 & 0.544 & 0.512 & 0.614 & 0.562 & 6.2 & 0.551 \\
 Entropy  & 0.523 & 0.540 & 0.514 & 0.571 & 0.559 & 6.8 & 0.541 \\
 \midrule
  \multicolumn{8}{c}{OLMoE-1B-7B} \\
\midrule
 IE (LR) & 0.732 & 0.737 & 0.830 & \textbf{0.787} & 0.726 & \textbf{2.0} & \textbf{0.762} \\
 IE (MLP) & \textbf{0.740} & \textbf{0.741} & \textbf{0.840} & 0.742 & 0.704 & \textbf{2.0} & 0.754 \\
 IE (RF) & 0.618 & 0.531 & 0.633 & 0.483 & 0.707 & 5.6 & 0.594 \\
 IE (Transf.) & 0.610 & 0.568 & 0.718 & 0.420 & \textbf{0.777} & 5.0 & 0.619 \\
 IE (XGB) & 0.739 & 0.728 & 0.810 & 0.682 & 0.719 & 2.8 & 0.736 \\
 LC (att.) & 0.258 & 0.287 & 0.267 & 0.481 & 0.224 & 7.8 & 0.303 \\
 LC (hid.) & 0.590 & 0.569 & 0.590 & 0.670 & 0.609 & 6.0 & 0.605 \\
 Entropy  & 0.636 & 0.621 & 0.653 & 0.650 & 0.634 & 4.8 & 0.639 \\
\bottomrule
\end{tabular}
\end{table}

Table~\ref{tab:token_level_results} reports token-level AUROC for \method and
all baselines that produce per-token scores, across five evaluation datasets
and both host models.\footnote{Sampling-based methods (SelfCheckGPT, Semantic
Uncertainty, Semantic Energy) and HaluNet operate at answer level by
construction and are therefore excluded from the token-level comparison.}

\paragraph{Predictive performance.} All five \method classifier variants exceed the best
single-signal baseline in average token-level AUROC on both host models, with
the strongest variant reaching 0.762 on OLMoE (LR) and 0.753 on
Gemma (XGBoost), improvements of 0.12 and 0.20 over the best baseline.
Classifier is not stable across host models: XGBoost and MLP lead on
Gemma, while LR and MLP perform best on OLMoE. Among single-signal baselines,
logit entropy and the LLM-Check hidden state score are the strongest; the
LLM-Check attention score yields AUROC below 0.5 on both models, as the
original LLM-Check formulation produces a single cumulative
score per answer and our per-token adaptation did not yield a reliable
per-token indicator. On the temporally out-of-distribution RealTimeQA test set,
\method achieves its highest per-dataset AUROC, suggesting the learned mapping
does not overfit to the training period.

\paragraph{Cross-dataset generalization.} \method maintains a consistent
advantage over single-signal baselines across all five evaluation datasets,
including the three out-of-distribution QA datasets (FreshQA, NQ-Open,
TruthfulQA) that differ substantially from the RealTimeQA training distribution
(SQuAD provides evidence in a similar way RealTimeQA does). Threshold-dependent
results (F1) are reported in Appendix~\ref{sec:per-token-cont}.

\subsection{Comparison to baselines (RQ2)}\label{subsec:comparison-baselines}

\begin{table}[htb]
\caption{Answer-level results (AUROC) across datasets and host models.}
\label{tab:answer_level_results}
\footnotesize
\setlength{\tabcolsep}{2pt}
\begin{tabular}{@{}lccccccc@{}}
\toprule
  & FQA & NQO & RTQA & SQuAD & TQA & Rank & Avg \\
\midrule
  \multicolumn{8}{c}{Gemma-4-26B} \\
\midrule
 HaluNet      & 0.836 & 0.835 & 0.937 & 0.945 & 0.952 & 3.2 & 0.901 \\
 IE (LR)      & 0.847 & 0.790 & 0.927 & 0.933 & 0.963 & 3.8 & 0.892 \\
 IE (MLP)     &  \textbf{0.866} & 0.780 & \textbf{0.939} & 0.943 & 0.952 & 3.0 & 0.896 \\ 
 IE (RF)      &  0.752 & 0.832 & 0.870 & 0.919 & 0.937 & 5.8 & 0.862 \\
 IE (Transf.) &  0.813 & 0.854 & 0.902 & 0.883 & 0.947 & 5.0 & 0.880 \\
 IE (XGB)     &  0.843 & \textbf{0.865} & 0.938 & 0.940 & \textbf{0.974} & \textbf{2.2} & \textbf{0.912} \\
 LC (att.)    &  0.650 & 0.711 & 0.616 & 0.887 & 0.724 & 8.0 & 0.718 \\
 LC (hid.)    & 0.680 & 0.834 & 0.445 & \textbf{0.949} & 0.704 & 7.0 & 0.722 \\
 Entropy      & 0.436 & 0.472 & 0.701 & 0.716 & 0.695 & 11.0 & 0.604 \\
 Perplexity   & 0.600 & 0.612 & 0.562 & 0.595 & 0.669 & 10.8 & 0.608 \\
 SCGPT (NLI)  & 0.646 & 0.630 & 0.713 & 0.760 & 0.788 & 8.2 & 0.707 \\
 SCGPT (P)    & 0.570 & 0.557 & 0.551 & 0.551 & 0.473 & 12.2 & 0.540 \\
 SemEnergy    & 0.437 & 0.429 & 0.553 & 0.506 & 0.452 & 13.4 & 0.475 \\
 SemUncert    & 0.526 & 0.512 & 0.614 & 0.670 & 0.597 & 11.4 & 0.584 \\

 \midrule
  \multicolumn{8}{c}{OLMoE-1B-7B} \\
\midrule
 HaluNet      & 0.804 & 0.842 & \textbf{0.953} & \textbf{0.881} & 0.778 & 4.6 & 0.851 \\
 IE (LR)      & 0.920 & 0.784 & 0.924 & 0.802 & 0.898 & 4.2 & 0.865 \\
 IE (MLP)     & \textbf{0.973} & 0.923 & \textbf{0.953} & 0.692 & 0.826 & 4.2 & 0.874 \\
 IE (RF)      & 0.790 & 0.821 & 0.565 & 0.714 & 0.776 & 8.6 & 0.733 \\
 IE (Transf.) & 0.721 & 0.510 & 0.649 & 0.329 & 0.839 & 10.0 & 0.610 \\
 IE (XGB)     & 0.943 & \textbf{0.926} & 0.928 & 0.786 & 0.825 & 4.0 & \textbf{0.882} \\
 LC (att.)    & 0.631 & 0.737 & 0.614 & 0.805 & 0.385 & 9.2 & 0.634 \\
 LC (hid.)    & 0.856 & 0.854 & 0.721 & 0.813 & 0.838 & 5.0 & 0.817 \\
 Entropy      & 0.245 & 0.170 & 0.530 & 0.768 & 0.191 & 11.8 & 0.381 \\
 Perplexity   & 0.146 & 0.051 & 0.182 & 0.367 & 0.113 & 13.8 & 0.172 \\
 SCGPT (NLI)  & 0.851 & 0.685 & 0.812 & 0.706 & 0.869 & 7.0 & 0.784 \\
 SCGPT (P)    & 0.622 & 0.550 & 0.644 & 0.564 & 0.531 & 11.2 & 0.582 \\
 SemEnergy    & 0.866 & 0.805 & 0.910 & 0.840 & \textbf{0.907} & \textbf{3.6} & 0.866 \\
 SemUncert    & 0.768 & 0.713 & 0.803 & 0.688 & 0.877 & 7.8 & 0.770 \\
\bottomrule
\end{tabular}
\end{table}

\begin{figure*}[htb]
  \centering
  \includegraphics[width=1\textwidth]{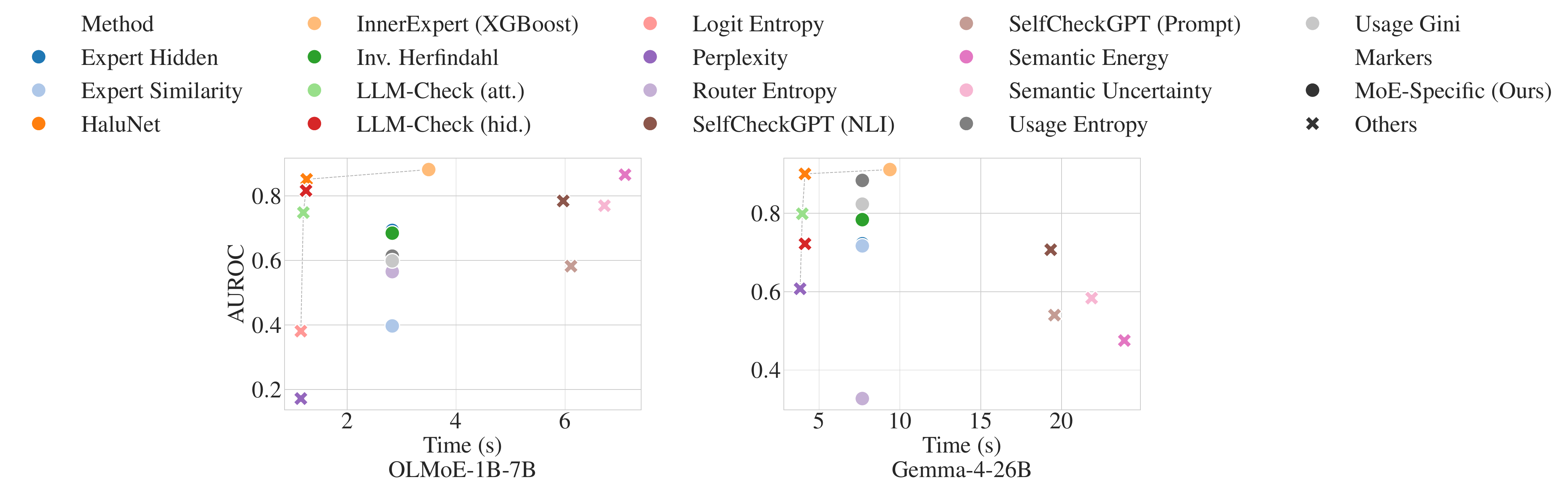}
  \caption{
    Inference time (per 100 tokens) vs.\ answer-level AUROC (averaged across
    five evaluation datasets) for each detection method, measured on 50
    RealTimeQA questions per host model. The gray, dashed line represents the
    pareto frontier. 
  }
  \label{fig:inference_time_vs_auroc}
\end{figure*}

Table~\ref{tab:answer_level_results} reports answer-level AUROC for \method
against all baselines.

\paragraph{Baselines.} \method (XGB) achieves the best average AUROC
on both host models, surpassing
HaluNet, the strongest trainable baseline, despite using
structurally different input features and a complex classifier architecture.
Sampling-based
methods exhibit the largest cross-model variance: Semantic Energy is
competitive on OLMoE (0.866) but near-random on Gemma (0.475), and Semantic
Uncertainty shows a similar instability; SelfCheckGPT (NLI) is the most stable
sampling-based baseline. Among internal-signal baselines,
the LLM-Check hidden state score achieves competitive results.
Threshold-dependent results (F1) are reported in
Appendix~\ref{sec:comparison-baselines-cont}.

\paragraph{Inference cost.} Figure~\ref{fig:inference_time_vs_auroc} shows,
\method (XGB) reaches the Pareto frontier, exceeding the AUROC of the strongest
baselines while remaining single-pass, whereas sampling-based methods cluster
in the high-cost region without a proportionate predictive performance gain. We
provide a complete table with the inference times and peak memory usage in
Appendix~\ref{sec:comparison-baselines-cont},
Table~\ref{tab:inference_benchmark}.
Vanilla generation costs 1.15\,s per 100 tokens
on OLMoE and 3.83\,s on Gemma. Standard internal-signal baselines
add negligible overhead ($<\!10\%$). Extracting the full set of MoE signals
raises the cost to $\sim\!2.5\times$ vanilla, and the \method classifier adds
only $\sim\!0.7$\,s on top ($\sim\!3\times$ vanilla overall), with negligible
difference across classifier variants and only $\sim\!3\%$ peak GPU memory
overhead. HaluNet is substantially cheaper but underperforms \method in
AUROC. Sampling-based methods are the most expensive category due to
$K{=}5$ generation passes, with SelfCheckGPT (Prompt) requiring a second
model instance that roughly doubles GPU memory.

\subsection{Signal contribution analysis (RQ3)}\label{subsec:signals-analysis}

Table~\ref{tab:signal_contribution_AUROC} isolates the contribution of
individual MoE-specific signals by reporting answer- and token-level AUROC
for each signal independently, alongside the standard signals and the full
\method classifier.

\begin{table}[htb]
\caption{Average AUROC for individual MoE signals, baselines, 
  \method (XGB). Each value is averaged across all five evaluation datasets
  per host model. Answer-level scores use the best aggregation per method. Methods
  without a token-level counterpart are marked ``---''.}
\label{tab:signal_contribution_AUROC}
\centering
\footnotesize
\setlength{\tabcolsep}{3pt}
\begin{tabular}{lcccc}
\toprule
 & \multicolumn{2}{r}{OLMoE-1B-7B} & \multicolumn{2}{r}{Gemma-4-26B} \\
  Method & Answer & Token & Answer & Token \\
\midrule
  Inv. Herfindahl   & \textbf{0.685} & \textbf{0.675} & 0.784 & 0.602 \\
  Exp. Entropy      & 0.613 & 0.574 & \textbf{0.884} & 0.591 \\
  Exp. Gini         & 0.599 & 0.561 & 0.824 & \textbf{0.626} \\
  Expert Hidden     & 0.693 & 0.640 & 0.722 & 0.466 \\
  Expert Similarity & 0.397 & 0.442 & 0.717 & 0.437 \\
  Router Entropy    & 0.565 & 0.574 & 0.327 & 0.446 \\
\midrule
  Hid. Score        & \textbf{0.817} & 0.605 & 0.722 & \textbf{0.551} \\
  Att. Score        & 0.748 & 0.303 & \textbf{0.799} & 0.313 \\
  Logit Entropy     & 0.381 & \textbf{0.639} & 0.604 & 0.541 \\
  Perplexity        & 0.172 & --- & 0.608 & --- \\
\midrule
  IE (XGB)          & \textbf{0.882} & 0.736 & \textbf{0.912} & \textbf{0.753} \\
  IE (MLP)          & 0.874 & \textbf{0.754} & 0.896 & 0.730 \\
  HaluNet           & 0.851 & --- & 0.901 & --- \\
  SCGPT (NLI)       & 0.784 & --- & 0.707 & --- \\
  SCGPT (Prompt)    & 0.582 & --- & 0.540 & --- \\
  SemUncert         & 0.770 & --- & 0.584 & --- \\
  SemEnergy         & 0.866 & --- & 0.475 & --- \\
\bottomrule
\end{tabular}
\end{table}

\paragraph{Individual MoE signals.} No single MoE signal is uniformly strong
across both host models. Expert entropy and usage Gini are the strongest
individual MoE signals on Gemma,
approaching HaluNet, but are considerably weaker on OLMoE. Conversely,
expert hidden score is the strongest
individual MoE signal on OLMoE but is mid-ranking on Gemma.
Expert similarity is near-random on OLMoE yet moderately informative
on Gemma, and router entropy is anti-correlated on Gemma.

\paragraph{Combination benefit.} Combining all signals via the \method
classifier yields consistent gains over the best individual signal, but the
magnitude depends on the host model. On OLMoE, where individual MoE
signals are weak, the combination provides a significant improvement: IE (XGBoost)
reaches 0.882 answer-level AUROC, a gain of 0.197 over the best individual
MoE signal. On Gemma, where expert entropy alone already achieves
0.884, the combination adds a consistent margin (IE
XGBoost: 0.912). At the token level, the signal combination yields a larger
relative gain. These results
indicate that the \method classifier extracts complementary information from
the signal set that is not available to any individual signal.

\paragraph{Standard signals.} Among the standard signals, the LLM-Check
hidden score is strong on OLMoE (0.817 answer) but drops on Gemma (0.722),
while the attention score shows the opposite pattern (0.748 vs.\ 0.799) but
degrades to near-random at the token level. Logit entropy is
anti-correlated at answer level on OLMoE but moderately informative
at the token level (0.639). Threshold-dependent results (F1) are reported in
Appendix~\ref{sec:signal-analysis-cont}.

\section{Conclusion}\label{sec:conclusion}

We presented \method, a single-pass, per-token hallucination detector that
leverages MoE-specific internal signals, alongside standard transformer
signals. Extensive experiments show that \method achieves up to 0.91
answer-level and 0.76 token-level AUROC, outperforming baselines spanning the
sampling-based, internal-signal, and trainable-detector paradigms, while
requiring only a single forward pass and modest computational overhead. Our
signal contribution analysis reveals that no individual MoE signal is reliably
strong across architectures, but combining them yields consistent gains,
demonstrating that the combination of these signals yields information
unavailable to any individual signal. 

\paragraph{Limitations.} Our human validation study
(Appendix~\ref{subsec:llm-judge-validation}) was evaluated on a small scale; we
aim to expand it in future work with a larger sample size and annotator
diversity. Additionally, our current evaluation is limited to the English
language; future work will explore the generalizability of our approach over a
more diverse set of languages. Finally, this paper demonstrates that
MoE-specific signals can be leveraged for hallucination detection. Hence, the
classifiers used in \method are naturally lightweight. See
appendix~\ref{sec:future-work} for details.

\section*{Acknowledgments}

Work supported by national funds through Fundação para a Ciência e a Tecnologia, I.P. (FCT) under projects: UID/50021/2025 ({\small \url{https://doi.org/10.54499/UID/50021/2025}}), UID/PRR/50021/2025 ({\small \url{https://doi.org/10.54499/UID/PRR/50021/2025}}), and  CCloud (ref. 2023.16986.ICDT, {\small \url{https://doi.org/10.54499/2023.16986.ICDT}}). This work was also supported by the European Union’s Horizon Europe research and innovation program under Grant Agreement GAP-101189689.

\bibliography{references}

\clearpage

\appendix

\section{Hallucination Detection via Epistemic Uncertainty}\label{sec:epistemic-uncertainty}

Uncertainty in next-token prediction can be decomposed into \emph{aleatoric}
and \emph{epistemic} uncertainty~\cite{hullermeier2021aleatoric}.
\emph{Aleatoric} uncertainty reflects irreducible ambiguity in the data
generating process (\eg multiple valid continuations of a prefix such as ``The
weather is \ldots''), and is directly reflected in the entropy of the model's
output distribution. \emph{Epistemic} uncertainty arises from the model's lack
of knowledge about the input region and is reducible with additional evidence.
Existing hallucination detectors tend to operate on signals that primarily
reflect aleatoric or aggregate uncertainty (\eg via multiple stochastic
generations or logit-based predictions), leaving epistemic
uncertainty largely under-explored as a detection signal.

The connection between epistemic uncertainty and hallucination provides a
theoretical hook for detection: when epistemic uncertainty is high but the
model emits a confident-looking answer (regardless of its aleatoric
uncertainty), the output is likely to be unreliable and therefore
hallucinated~\cite{yadkori2024believe}. This suggests that practical proxies
for epistemic uncertainty can serve as useful hallucination detectors. The
contributions in this paper allow defining useful epistemic uncertainty proxies
for hallucination detection that are cheap to compute and available within a
single forward pass, without requiring multiple stochastic generations or an
external verifier. In the vision domain, MoE-specific signals have been shown
to capture epistemic uncertainty, with expert disagreement acting as an
analogue of ensemble disagreement~\cite{pavlitska2025extracting}. 

\textbf{Uncertainty Decomposition.}
Following the Bayesian treatment of predictive
uncertainty~\cite{hullermeier2021aleatoric}, suppose the model parameters lie
in a posterior $p(\theta)$.
The marginal predictive distribution averages over this posterior,
\begin{equation}\label{eq:marginal}
  p(y_t\mid \mathbf{x}, y_{<t}) = \mathbb{E}_{p(\theta)}\!\left[\, p(y_t\mid \mathbf{x}, y_{<t}, \theta)\,\right].
\end{equation}
The total predictive uncertainty, measured as the Shannon entropy of the
marginal predictive, decomposes as
\begin{equation}\label{eq:decomp}
\begin{split}
  \underbrace{\mathrm{H}\!\left[p(y_t\mid \mathbf{x}, y_{<t})\right]}_{\text{total}}
  &= \underbrace{
      \mathbb{E}_{p(\theta)}\!\left[
        \mathrm{H}\!\left[p(y_t\mid \mathbf{x}, y_{<t}, \theta)\right]\right]   
      }_{\text{aleatoric}} \\
  &\quad + \underbrace{
      \mathrm{I}(y_t;\,\theta\mid \mathbf{x}, y_{<t})
    }_{\text{epistemic}},
\end{split}
\end{equation}
where $\mathrm{I}(\cdot;\cdot\mid\cdot)$ is the conditional mutual information between
$y_t$ and $\theta$~\cite{smith2025rethinking, gal2017uncertainty}. The
\emph{aleatoric} component
captures irreducible ambiguity in the data-generating process (\eg multiple
valid continuations of a prefix such as ``The weather is \ldots'') and is
reflected in the entropy of a single conditional distribution
$p(y_t\mid \mathbf{x}, y_{<t}, \theta)$. The \emph{epistemic} component
captures uncertainty about the parameters themselves and is
reducible with additional evidence. Since $p(\theta)$ is not
available in LLMs, neither term of \eqref{eq:decomp} can be computed exactly; in
practice, estimating uncertainty involves multiple model evaluations
or stochastic generations, which are expensive at inference time. This is the
case for methods like Semantic Uncertainty~\cite{kuhn2023semantic} and SelfCheckGPT~\cite{manakul2023selfcheckgpt}, which
estimate an aggregate uncertainty measure that mixes aleatoric and epistemic.

\textbf{Epistemic Uncertainty and Hallucination.}
In LLMs, the decomposition in \eqref{eq:decomp} links epistemic uncertainty and
hallucination~\cite{yadkori2024believe}: when epistemic uncertainty is high
relative to aleatoric uncertainty, the model's output is likely to be
hallucinated. The mutual information can be
expressed as an expected divergence
among posterior samples,
\begin{equation}\label{eq:epistemic-kl}
\begin{split}
  \mathrm{I}(y_t;\,\theta\mid \mathbf{x}, &y_{<t})
  = \mathbb{E}_{p(\theta)}\!\Big[ \\
    &\quad \mathrm{KL}\!\left(
      p_\theta(y_t\mid \mathbf{x}, y_{<t}) \,\big\|\, p(y_t\mid \mathbf{x}, y_{<t})
    \right)\Big],
\end{split}
\end{equation}
where $\mathrm{KL}(\cdot\|\cdot)$ is the Kullback-Leibler divergence. A token
$y_t$ is therefore likely to be hallucinated when this expected divergence is
high relative to the aleatoric term, \ie when the model's parameters, were they
resampled, would diverge substantially yet the model nonetheless emits a
confident-looking prediction. In the well-specified Bayesian limit this
within-model disagreement tracks divergence from the data-generating
distribution $p_{\text{data}}$, but this link is broken for misspecified
models such as LLMs, whose parameters do not admit a posterior $p(\theta)$. We therefore treat MoE routing signals as empirical
proxies of within-model disagreement, following the empirical results of
\citet{pavlitska2025extracting} that expert disagreement captures epistemic
uncertainty in vision models as the bridge. The challenge addressed in this
paper is to define such proxies that are available within a single forward pass
and do not require multiple stochastic generations or an external verifier.

\section{Related Work}\label{sec:related-work}

We review prior work along the three paradigms introduced in
Section~\ref{sec:introduction}: sampling-based detection, internal-signal
detection, and trainable detectors. We then discuss uncertainty estimation
methods, which provide the theoretical motivation for \method, and conclude
with the emerging literature on MoE-specific signals, which constitutes the gap
our work addresses.

\paragraph{Sampling-based approaches} probe the consistency of a model's
outputs across multiple stochastic generations. \citet{manakul2023selfcheckgpt}
introduce SelfCheckGPT, which detects hallucinations by measuring agreement
among several sampled responses to the same prompt, either via an LLM-based
checker or through token-level n-gram consistency. In this work, we use the two
best-performing SelfCheckGPT variants for comparison: an NLI-based variant that
detects contradictions between the original response and sampled responses
using a DeBERTa model fine-tuned on MNLI, and a prompt-based variant that uses
an LLM to assess whether each response sentence is supported by the sampled
responses. \citet{kuhn2023semantic} propose Semantic Uncertainty, which
clusters semantically equivalent samples and computes entropy over the
resulting clusters, providing a more robust uncertainty estimate than raw
token-level entropy. \citet{farquhar2024detecting} scale this approach and
demonstrate that semantic entropy is effective for hallucination detection
across long-form generation tasks. \citet{ma2025semantic} extend this line of
work with Semantic Energy, which replaces the probability-space entropy of
Semantic Uncertainty with a Boltzmann-inspired energy function over semantic
clusters, capturing model confidence even when sampled responses are
semantically identical. \citet{chen2024inside} introduce INSIDE, which measures the semantic
consistency of multiple sampled responses in the model's internal embedding
space. Their EigenScore computes the log-determinant of the covariance matrix
of sentence embeddings across $K$ sampled responses, capturing semantic
divergence in the dense representation space. While this approach retains more
semantic information than text-level consistency metrics, it still requires
multiple stochastic generations per prompt. \citet{han2024semantic} introduce
Semantic Entropy Probes, lightweight probes trained to approximate semantic
entropy from internal states, bridging sampling-based and internal-signal
approaches. While these methods are effective, they require multiple forward
passes per prompt, making them prohibitively expensive at inference time. \method, by contrast,
operates on signals extracted from a single forward pass and requires no
additional generations.

\paragraph{Internal signal-based methods} leverage internal model states (\eg
hidden representations, attention patterns, or output distributions) collected
during a single forward pass. \citet{azaria2023internal} show that a classifier
trained on hidden states can distinguish factual from non-factual statements,
demonstrating that hallucination-relevant information is encoded in internal
representations. \citet{sriramanan2024llm} propose LLM-Check, which compute
hallucination scores from hidden states, attention kernels, and perplexity.
These methods share \method's single-pass efficiency, but they
operate exclusively on individual signals available in any transformer
architecture (hidden states, attention matrices, and output logits) and do not
exploit the routing structure unique to MoE models. \method extends this
paradigm by introducing MoE-specific signals (\eg router entropy, expert
disagreement, expert usage distributions) that are absent in dense
architectures while combining them into a single hallucination score.

A related but distinct line of work modifies internal representations at
inference time to \emph{mitigate} hallucinations rather than detect them.
Decoding by Contrasting Layers~\cite{chuang2024dola} contrasts logits
across layers to amplify factual knowledge localized in specific
transformer layers. Inference-Time Intervention~\cite{li2023inference}
shifts activations along truth-correlated directions identified in a
sparse set of attention heads. TruthX~\cite{zhang2024truthx} edits hidden
representations in a learned truthful space to improve output factuality.
MoLE~\cite{liang2025mole} extends the layer-contrasting idea to
vision-language models, dynamically selecting transformer layers as
``experts'' (Final, Second Opinion, and Prompt Retention) via a heuristic
gating mechanism. We note that MoLE's use of ``expert'' refers to
transformer layers, not the routed feedforward networks of MoE
architectures. These methods are complementary to hallucination detection
but fall outside the scope of this paper, which focuses on detection
rather than mitigation.

\paragraph{Trainable detectors over internal signals} typically use lightweight
classifiers over internal signals to produce hallucination scores.
\citet{tong2025halunet} introduce HaluNet, which models hallucination risk
using a multi-branch architecture that fuses token-level log-likelihoods,
entropy scores, and hidden-state embeddings. \citet{wang2026joint} propose a
joint evaluation framework that assesses both answer and reasoning consistency
for hallucination detection in reasoning models. \citet{wang2025faclens}
present FacLens, a transferable probe for non-factuality prediction that
analyzes how factual knowledge is embedded in hidden representations of
fact-seeking questions, with the goal of cross-model transferability.
\citet{su2024unsupervised} present MIND, a framework that trains an MLP on
hidden states collected during generation, using an automated Wikipedia-based
labeling scheme to avoid manual annotation. \method follows this
trainable-detector paradigm but differs in two key respects: (i) it enriches
the feature set with MoE-specific routing signals unavailable to prior methods,
and (ii) it employs an unsupervised labeling pipeline based on LLM-as-judge
evaluation against reference evidence, requiring no human annotations.

\paragraph{Uncertainty estimation in LLMs.} Uncertainty estimation provides the
theoretical motivation for \method, developed in
Appendix~\ref{sec:epistemic-uncertainty}. Generally, for Machine Learning,
\citet{hullermeier2021aleatoric} provide a formalization of aleatoric
uncertainty (irreducible ambiguity in the data-generating process) and
epistemic uncertainty (uncertainty arising from the model's lack of knowledge),
which is reducible with additional evidence. \citet{lahlou2023deup} propose
DEUP, a framework for directly predicting epistemic uncertainty by learning to
estimate excess risk and subtracting an estimate of aleatoric uncertainty.
\citet{smith2025rethinking} revisit the aleatoric/epistemic decomposition and
highlight practical challenges in disentangling the two components in deep
learning models. Specifically for LLMs, \citet{yadkori2024believe} introduce an
iterative prompting approach for estimating epistemic uncertainty in LLMs via
iterative prompting based on previous responses. \citet{xia2025survey} provide
a comprehensive survey of uncertainty estimation methods for LLMs, covering
both sampling-based and internal-signal approaches. 

\paragraph{Signals from Mixture-of-Experts} The MoE
architecture~\cite{fedus2022switch} introduces a routing mechanism that selects
a sparse subset of expert networks per token, producing signals (router
distributions, per-expert hidden states and expert usage patterns) that are not
observable in dense models. Despite the growing prominence of MoE models in the
LLM landscape, the information encoded in these signals remains largely
unexplored for hallucination detection. \citet{pavlitska2025extracting}
demonstrate that uncertainty estimates can be extracted from MoE models in the
vision domain, showing that expert disagreement serves as an analogue of
ensemble disagreement and captures epistemic uncertainty in semantic
segmentation tasks. To the best of our knowledge, no prior work leverages
MoE-specific internal signals for hallucination detection in LLMs.

\section{Additional Method Details}\label{sec:method-details}

This appendix provides a complete reference for the signals used by \method
(Table~\ref{tab:signal-summary}), the inference procedure
(Algorithm~\ref{alg:inference}), and the training pipeline
(Algorithm~\ref{alg:training}).

\subsection{Signal Summary}

\begin{table*}[htb]
  \caption{
    Summary of signals extracted by \method. Each signal is computed cumulatively
    over the generated prefix up to position $t$, yielding a per-token value. The
    total per-token feature dimensionality is the sum of all signal dimensions.
    $L$ = number of MoE layers, $H$ = number of attention heads, $N$ = number of
    experts.
  }
  \label{tab:signal-summary}
  \centering
  \footnotesize
  \setlength{\tabcolsep}{4pt}
  \begin{tabular}{@{}lllllr@{}}
    \toprule
    Signal & Notation & Category & Description & Per-token dim. \\
    \midrule
    Hidden state score & $\phi^{\mathrm{hid}}_{t,l}$ & Standard &
    Mean log-singular-value of hidden state covariance & $L$ \\
    Attention score & $\phi^{\mathrm{att}}_{t,l,h}$ & Standard &
    Log-cumulative sum of attention kernel diagonal & $LH$ \\
    \midrule
    Router entropy & $\phi^{\mathrm{rout}}_{t,l}$ & MoE &
    Shannon entropy of gating distribution & $L$ \\
    Expert hidden score & $\phi^{\mathrm{exp\text{-}hid}}_{t,l}$ & MoE &
    Routing-weighted hidden score of selected experts & $L$ \\
    Expert similarity & $\phi^{\mathrm{sim}}_{t,l}$ & MoE &
    Routing-weighted pairwise cosine similarity of experts & $L$ \\
    Expert usage & $\mathbf{u}_{t,l}$ & MoE &
    Cumulative routing-weighted usage distribution & $LN$ \\
    Gini impurity & $\phi^{\mathrm{gini}}_{t,l}$ & MoE &
    Gini impurity of usage distribution & $L$ \\
    Inverse Herfindahl & $\phi^{\mathrm{herf}}_{t,l}$ & MoE &
    Effective number of experts from usage distribution & $L$ \\
    \midrule
    \multicolumn{4}{@{}l}{\textbf{Total per-token dimensionality}}
    & $5L{+}LH{+}LN$ \\
    \bottomrule
  \end{tabular}
\end{table*}

Table~\ref{tab:signal-summary} lists all signals extracted by \method, their
notation, category, and per-token dimensionality. Signals are grouped into
\emph{standard} signals (available in any transformer) and \emph{MoE-specific}
signals (derived from the MoE block). All per-layer signals are computed
cumulatively over the generated prefix $y_{\leq t}$, yielding a per-token value
at each generation step $t$. The ratio between the total per-token
dimensionality and the original dimensionality of the signals used from the
host model underscores the compactness of the representation, \ie detecting
hallucinations does not require the full internal state of the host model. For
the two host models used in this paper, the aggregate feature dimensionalities
are presented in Table~\ref{tab:feature-dimensions}.

\begin{table}[H]
  \caption{
    Feature dimensionality for each host model. $L$: number of layers, $H$:
    number of attention heads, $N$: number of experts, $|\Phi_t|$: per-token
    feature dimensionality, Params: total model parameters, Act.: number of
    active parameters per token.
  }
  \label{tab:feature-dimensions}
  \centering
  \footnotesize
  \setlength{\tabcolsep}{3pt}
  \begin{tabular}{@{}lrrrrrr@{}}
    \toprule
    Model & $L$ & $H$ & $N$ & $|\Phi_t|$ & Params & Act. \\
    \midrule
    OLMoE-1B-7B-0924-Instruct & 16 & 16 & 64 & 1\,360 & $\sim$7B & $\sim$1B\\
    Gemma-4-26B-A4B-it & 30 & 16 & 129 & 4\,470 & $\sim$26B & $\sim$4B \\
    \bottomrule
  \end{tabular}
\end{table}

For Gemma, $N{=}129$ comprises 128 routed experts plus 1 shared expert
that is always active.

\subsection{Inference Procedure}

Algorithm~\ref{alg:inference} outlines the \method inference pipeline. Given a
question $\mathbf{x}$ (and optional evidence $\mathbf{e}$), the host model
generates an answer $\mathbf{y} = (y_1, \dots, y_T)$ via greedy decoding.
During generation, internal signals are collected at each token position and
assembled into the feature vector $\Phi_t$ (Eq.~\ref{eq:feature-vector}). The
trained classifier $s(\cdot)$ produces a per-token hallucination score
$s(\Phi_t) \in [0,1]$. The answer-level score $S(\mathbf{y})$ is obtained by
aggregating per-token scores via the max operator (Eq.~\ref{eq:answer-score}).

\begin{algorithm}[t]
\caption{\method Inference}\label{alg:inference}
\begin{algorithmic}[1]
\REQUIRE Question $\mathbf{x}$, evidence $\mathbf{e}$ (optional), trained classifier $s$, host model $\theta$
\ENSURE Answer $\mathbf{y}$, per-token scores $\{s(\Phi_t)\}_{t=1}^T$, answer-level score $S(\mathbf{y})$
\STATE Generate answer $\mathbf{y} = (y_1, \dots, y_T)$ via greedy decoding on $\theta$, collecting internal states at each step
\FOR{$t = 1$ \TO $T$}
  \FOR{$l = 1$ \TO $L$}
    \STATE Collect hidden states $\mathbf{H}_{t,l} = \{\mathbf{h}_{1,l}, \dots, \mathbf{h}_{t,l}\}$
    \STATE Compute $\phi^{\mathrm{hid}}_{t,l}$ \hfill $\triangleright$ Eq.~\ref{eq:hidden-score}
    \FOR{$h = 1$ \TO $H$}
      \STATE Compute $\phi^{\mathrm{att}}_{t,l,h}$ \hfill $\triangleright$ Eq.~\ref{eq:attention-score}
    \ENDFOR
    \STATE Compute $\phi^{\mathrm{rout}}_{t,l}$ \hfill $\triangleright$ Eq.~\ref{eq:router-entropy}
    \STATE Compute $\phi^{\mathrm{exp\text{-}hid}}_{t,l}$ \hfill $\triangleright$ Eq.~\ref{eq:expert-hidden-score}
    \STATE Compute $\phi^{\mathrm{sim}}_{t,l}$ \hfill $\triangleright$ Eq.~\ref{eq:expert-similarity}
    \STATE Update $\mathbf{u}_{t,l}$ \hfill $\triangleright$ Eq.~\ref{eq:expert-usage}
    \STATE Compute $\phi^{\mathrm{gini}}_{t,l}$, $\phi^{\mathrm{herf}}_{t,l}$ \hfill $\triangleright$ Eqs.~\ref{eq:gini}, \ref{eq:herfindahl}
  \ENDFOR
  \STATE Assemble $\Phi_t$ \hfill $\triangleright$ Eq.~\ref{eq:feature-vector}
  \STATE Compute $s(\Phi_t) \in [0,1]$
\ENDFOR
\STATE Aggregate $S(\mathbf{y}) = \max_{t=1}^T s(\Phi_t)$ \hfill $\triangleright$ Eq.~\ref{eq:answer-score}
\RETURN $\mathbf{y}$, $\{s(\Phi_t)\}_{t=1}^T$, $S(\mathbf{y})$
\end{algorithmic}
\end{algorithm}

\subsection{Training Procedure}

Algorithm~\ref{alg:training} outlines the \method training pipeline. Training
data is sourced from RealtimeQA (January 2024--December 2025), with answers
generated under two conditions (base and RAG) per question. The LLM-as-a-judge
produces answer-level hallucination labels and hallucinated span strings, which
are mapped to token-level binary labels. Feature vectors are extracted for all
generated tokens and assembled into a training matrix. A stratified
train/validation split, grouped by question identifier, prevents leakage
between answers to the same question. Grid search over five classifier families
(Logistic Regression, Random Forest, XGBoost, MLP, and a Transformer encoder) is
scored by F1 on the validation fold. The best model per family is refit on the
combined train+validation data, and the decision threshold $\tau$ is
F1-optimized on this combined set via a precision-recall curve sweep.

\begin{algorithm}[t]
\caption{\method Training}\label{alg:training}
\begin{algorithmic}[1]
\REQUIRE RealtimeQA questions $\{\mathbf{x}_i\}$ with evidence $\{\mathbf{e}_i\}$, host model $\theta$, LLM judge $J$
\ENSURE Trained classifier $s$, threshold $\tau$
\STATE \textbf{Phase 1: Answer Generation \& Labeling}
\FOR{each question $\mathbf{x}_i$}
  \STATE Generate base answer $\mathbf{y}_i^{\mathrm{base}}$ (no evidence) via greedy decoding on $\theta$
  \IF{evidence $\mathbf{e}_i$ available}
    \STATE Generate RAG answer $\mathbf{y}_i^{\mathrm{rag}}$ (with evidence) via greedy decoding on $\theta$
  \ENDIF
\ENDFOR
\FOR{each generated answer $\mathbf{y}_i$}
  \STATE LLM judge $J$ produces binary label $\ell_i$ and hallucinated spans
  \STATE Map spans to token-level labels $\{\ell_{t}\}_{t=1}^{T_i}$
\ENDFOR
\STATE \textbf{Phase 2: Feature Extraction}
\FOR{each generated answer $\mathbf{y}_i$}
  \FOR{$t = 1$ \TO $T_i$}
    \STATE Extract $\Phi_t$ following Algorithm~\ref{alg:inference}, steps 2--11
  \ENDFOR
\ENDFOR
\STATE \textbf{Phase 3: Model Selection}
\STATE Split data into train/validation sets, grouped by question identifier
\STATE Grid search over classifier families $\{$LR, RF, XGBoost, MLP, Transformer$\}$, F1-scored on validation fold
\STATE Select best hyperparameters per family; refit on combined train+validation
\STATE Optimize threshold $\tau$ for F1 on combined set via PR curve sweep
\RETURN $s$, $\tau$
\end{algorithmic}
\end{algorithm}

\section{Implementation Details}\label{sec:implementation-details}

\paragraph{Expert hidden state extraction.}
Standard MoE implementations such as HuggingFace's \texttt{OlmoeSparseMoeBlock}
and \texttt{Gemma4TextExperts} do not expose per-expert hidden states---they
return only the routing-weighted combination of selected expert outputs. To
collect the per-expert signals required by \method (e.g., expert hidden score,
expert similarity), we replace (via monkey-patching) the \texttt{forward} method of each MoE block in
the host model with a modified version that saves the pre-routing per-expert
outputs to a \texttt{last\_experts\_hidden} attribute \emph{before} the routing
weights are applied. For Gemma, the host decoder layer flattens
hidden states from $(B, S, H)$ to $(B{\times}S, H)$ before calling the MoE block;
we handle this by registering a pre-forward hook on the parent layer that
captures the original batch and sequence dimensions, enabling the patched
forward to recover the correct 4D output shape. The design is
architecture-agnostic: adding a new MoE model requires only implementing a
\texttt{forward\_<model>} function.

\paragraph{Memory usage.}
The instrumentation adds moderate GPU memory overhead. Benchmarking on 50
RealtimeQA questions, peak GPU memory for OLMoE increases from
12.7\,GB (vanilla generation) to 13.1\,GB with full \method
instrumentation, an overhead of approximately 3\%. For Gemma, the
increase is from 46.8\,GB to 48.1\,GB (2.8\%). The dominant memory cost of
\method is the per-expert hidden state tensor, which has shape
$(B, S, k, d)$ per MoE layer per generation step. On resource-constrained
environments with limited VRAM, these tensors must be moved to CPU between
generation steps to free memory for subsequent inference calls. On our
hardware (2$\times$ 96\,GB VRAM), tensors can be accumulated on GPU and
transferred in a single batch after generation completes. During baseline
fitting, unused keys (e.g., \texttt{expert\_usage} at 28\,GB
padded for Gemma) are filtered during data loading, substantially reducing
peak RAM, at the cost of additional CPU time for filtering.

\paragraph{Storage requirements.}
Total disk usage for the project is approximately 1.7\,TB, largely due to raw
batch files saved during generation. Pretrained models (host LLMs and baseline
models combined) account for 119\,GB, while trained \method detectors take up
311\,MB. The training data (RealtimeQA 2024--2025) requires 0.9\,TB of raw
files, the temporal OOD test set (RealtimeQA 2026) adds 0.17\,TB, and the OOS
datasets contribute 0.48\,TB. 

\paragraph{HaluNet re-implementation} As no open-source implementation of
HaluNet~\cite{tong2025halunet} is available, we re-implement it from the
architecture described in the paper. The model has three branches: a
log-likelihood branch and an entropy branch, each consisting of mean pooling
followed by a 2-layer MLP, and a hidden-state embedding branch with two 1D
convolutional layers (kernel size 3, padding 1) with ReLU activations and
adaptive average pooling. Branch outputs are fused via attention and projected
to a single logit. The hidden dimension is 128, the maximum sequence length is
50 (zero-padded and truncated), and dropout is 0.5. The model is trained with
Adam (learning rate $10^{-3}$), binary cross-entropy loss, for 20 epochs with a
batch size of 32. Training uses the same LLM-as-judge binary hallucination
labels as \method. The decision threshold is F1-optimized on the validation
set. The embedding dimension is model-dependent and determined from the data at
training time. All other hyperparameters follow the defaults specified in the
original paper.

\section{Datasets Description}\label{sec:datasets-description}

This section provides detailed descriptions of the five
datasets used in our experiments. RealtimeQA serves as both training data
and temporal out-of-distribution test set, while SQuAD, TruthfulQA,
NQ-Open, and FreshQA are used for cross-dataset generalization evaluation.
Summary statistics, including sample counts, hallucination rates, and average
answer lengths per host model, are reported in
Table~\ref{tab:dataset_statistics}.

\paragraph{RealtimeQA}~\cite{kasai2023realtime} is a dynamic question-answering
platform that releases questions about current real-world events on a regular
basis. Each question is accompanied by reference evidence in the form of
retrieved passages. We use questions from January 2024 to December 2025 as
training data and questions from January to June 2026 (365 questions) as the
temporally out-of-distribution test set, evaluating whether the detector
generalizes to post-training events.

\paragraph{SQuAD}~\cite{rajpurkar2016squad} is a reading comprehension dataset
comprising over 100{,}000 questions posed by crowdworkers on Wikipedia
articles. Each question is paired with a context passage (reference evidence)
from which the answer span is drawn. We use the validation split and randomly
sample 200 questions.

\paragraph{TruthfulQA}~\cite{lin2022truthfulqa} is a benchmark of 817 questions
across 38 categories designed to test whether language models avoid mimicking
human falsehoods and common misconceptions. The dataset does not provide
reference evidence, relying on the model's internal knowledge. We use the
train split to randomly sample 200 questions and use the provided answer as
evidence.

\paragraph{NQ-Open}~\cite{kwiatkowski2019natural,lee2019latent} is an
open-domain question-answering benchmark derived from Google's Natural
Questions by discarding the accompanying evidence documents and retaining only
questions with short answers. The dataset does not provide reference evidence,
as the open-domain setting requires systems to retrieve evidence independently.
We use the validation split to randomly sample 200 questions and use the
provided answer as evidence.

\paragraph{FreshQA}~\cite{vu2024freshllms} is a dynamic benchmark with
questions requiring fast-changing world knowledge, including questions with
false premises that need to be debunked. We use a snapshot from November 2025
to randomly sample 200 questions.

\begin{table}
\caption{
  Dataset statistics for all evaluation datasets and host models. Labels are
  from the LLM-as-a-judge approach (GLM-5.1). ``Ans.'' is the average number of
  words in the generated answer.
}
\label{tab:dataset_statistics}
\centering
\footnotesize
\setlength{\tabcolsep}{2pt}
\begin{tabular}{@{}lccccccc@{}}
\toprule
Dataset & N & Base & Eviden. & Halluc. & Grounded & Halluc.\,\% & Ans. \\
\midrule
  \multicolumn{8}{c}{Gemma-4-26B} \\
\midrule
SQuAD & 400 & 200 & 200 & 172 & 228 & 43.0 & 17.1 \\
TQA & 399 & 199 & 200 & 194 & 205 & 48.6 & 19.2 \\
NQO & 400 & 200 & 200 & 233 & 167 & 58.2 & 16.0 \\
FQA & 400 & 200 & 200 & 226 & 174 & 56.5 & 13.3 \\
RTQA & 730 & 365 & 365 & 397 & 333 & 54.4 & 15.3 \\
\midrule
  \multicolumn{8}{c}{OLMoE-1B-7B} \\
\midrule
SQuAD & 400 & 200 & 200 & 309 & 91 & 77.2 & 43.4 \\
TQA & 400 & 200 & 200 & 381 & 19 & 95.2 & 46.9 \\
NQO & 400 & 200 & 200 & 381 & 19 & 95.2 & 43.3 \\
FQA & 400 & 200 & 200 & 378 & 22 & 94.5 & 37.8 \\
RTQA & 730 & 365 & 365 & 635 & 95 & 87.0 & 43.8 \\
\bottomrule
\end{tabular}
\end{table}

\section{Additional Experiment Details}\label{sec:additional-experiment-details}

Table~\ref{tab:hyperparameter-grids} reports the hyperparameter grids used for
each classifier family.

\begin{table}
  \caption{Hyperparameter grids for each classifier family. Hyperparameters used for OLMoE are boldfaced, the ones used for Gemma are underlined.}
  \label{tab:hyperparameter-grids}
  \centering
  \footnotesize
  \setlength{\tabcolsep}{2pt}
  \begin{tabular}{@{}lcl@{}}
    \toprule
    Classifier & Hyperparameter & Grid \\
    \midrule
    \makecell{Logistic \\ Regression} & $C$ & \makecell{$\{0.01, \underline{0.1}, 1.0,$\\ $ \textbf{10.0}\}$} \\
    \midrule
    \multirow{3}{*}{\makecell{Random \\ Forest}} & n\_estimators & $\{\underline{300}, 500, \textbf{1000}\}$ \\
     & max\_depth & $\{3, 6, \underline{\textbf{10}}\}$ \\
     & min\_samples\_leaf & $\{\underline{\textbf{1}}, 5\}$ \\
    \midrule
    \multirow{3}{*}{XGBoost} & n\_estimators & $\{300, 500, \underline{\textbf{1000}}\}$ \\
     & max\_depth & $\{3, 6, \underline{\textbf{10}}\}$ \\
     & learning\_rate & $\{0.001, 0.01, \underline{\textbf{0.1}}\}$ \\
    \midrule
    \multirow{5}{*}{MLP} & hidden\_layer\_sizes & \makecell{$\{\textbf{(128,)}, \underline{(256,)},$ \\ $(512,), (1024,)\}$} \\
     & $\alpha$ & \makecell{$\{10^{-4}, 10^{-3},$ \\ $\textbf{10}^{-2}, \underline{10}^{-1}\}$} \\
     & learning\_rate\_init & $\{\underline{10}^{-3}, \textbf{10}^{-4}\}$ \\
     & early\_stopping & $\{\text{True}, \underline{\textbf{\text{False}}}\}$ \\
    \midrule
    \multirow{6}{*}{Transformer} & d\_model & $\{\textbf{256}, \underline{512}\}$ \\
     & n\_heads & $\{\underline{\textbf{4}}\}$ \\
     & n\_transformer\_layers & $\{1, \underline{2}, \textbf{3}\}$ \\
     & dropout & $\{\textbf{0.1}, \underline{0.3}\}$ \\
     & optimizer\_lr & $\{\underline{10}^{-3}, \textbf{10}^{-4}\}$ \\
     & max\_epochs & $\{\underline{\textbf{500}}\}$ \\
    \bottomrule
  \end{tabular}
\end{table}

\subsection{Training Dataset Analysis}

The block below contains the prompt template used to generate answers from the
host models. The prompts are designed to elicit responses from the host model
with and without access to reference evidence.

\begin{tcolorbox}[
  colback=blue!5, colframe=blue!50,
  title={\textbf{Generation Prompt}},
]
\small
\textbf{System message:}
\begin{alltt}\small
  You are a helpful assistant who 
  provides accurate and very concise 
  answers to questions about recent 
  events. Today is \{DATE\}.
\end{alltt}

\textbf{User message (base, no evidence):}
\begin{alltt}\small
  \{QUESTION\}
\end{alltt}

\textbf{User message (with evidence):}
\begin{alltt}\small
  Evidence: \{EVIDENCE\}

  Question: \{QUESTION\}
\end{alltt}
\end{tcolorbox}

The block below contains the prompt template used to label the training dataset
using the LLM-as-a-judge approach. The prompt is designed to elicit a binary
classification of the model's answer as either hallucinated or grounded, along
with the specific spans of text that are hallucinated.

\begin{tcolorbox}[
  colback=red!5, colframe=red!50,
  title={\textbf{Judge Prompt}},
]
\small
\textbf{System message:}
\begin{alltt}\small
  Return concise, valid JSON only.
\end{alltt}

\textbf{User message template:}
\begin{alltt}\small
  Given a question, evidence, and model 
  answer, decide whether the answer
  contains unsupported or hallucinated 
  content. Return valid JSON only with
  this exact schema:
    \{
      "label": 1 or 0, 
      "hallucinated\_spans": [
        "exact substring from answer", 
        ...
      ]
    \}

  Rules:
  - label = 1 means hallucinated/
    unsupported, label = 0 means 
    grounded/correct.
  - hallucinated\_spans must be 
    exact substrings from the answer.
  - if label = 0, hallucinated\_spans 
    must be an empty list.

  Question: \{QUESTION\}
  Evidence: \{EVIDENCE\}
  Answer: \{ANSWER\}

\end{alltt}
\end{tcolorbox} 

The following blocks contain training data examples of hallucinations and
non-hallucinations, wherein the hallucinated spans (as assessed by the judge
LLM) are highlighted in red. 

\begin{tcolorbox}[
  colback=gray!8, colframe=black!60,
  title={\textbf{Example 1} \small (2024/01/05)},
]
\small
\textbf{Question:} Who was named Director of the Year at this week's Palm
  Springs International Film Festival Awards?

\textbf{Evidence:} The "Barbie" director was presented with the Director of the
  Year award by her film's stars Margot Robbie and America Ferrera at the
  awards ceremony in Palm Springs, California. Other winners included actors
  Cillian Murphy, Emma Stone and Carey Mulligan.                                                                                                                                                   

\textbf{OLMoE (no evidence)} \hfill \textsc{hallucinated}
\begin{alltt}\small
  At this week's Palm Springs 
  International Film Festival Awards, 
  the Director of the Year award was 
  presented to \textcolor{red}{Alfonso Cuarón, a 
  renowned Mexican film director, 
  producer, and screenwriter. Cuarón 
  is known for his work on films such 
  as "Y Tu Mamá También," " Children 
  of Men,"}
\end{alltt}

\textbf{OLMoE (with evidence)} \hfill \textsc{hallucinated}
\begin{alltt}\small
  The Director of the Year award was 
  presented to the "Barbie" film's 
  director, \textcolor{red}{Liesl Tommy}, at the Palm 
  Springs International Film Festival 
  Awards in California. Other winners 
  included actors Cillian Murphy, Emma 
  Stone, and Carey Mulligan.
\end{alltt}

\textbf{Gemma 4 (no evidence)} \hfill \textsc{hallucinated}
\begin{alltt}\small
  \textcolor{red}{No awards were presented at this 
  week's Palm Springs International Film 
  Festival, as the festival is 
  currently ongoing (January 4–14, 2024
  ). The awards ceremony is scheduled 
  for January 14, 2024.}
\end{alltt}

\textbf{Gemma 4 (with evidence)} \hfill \textsc{hallucinated}
\begin{alltt}\small
  \textcolor{red}{Greta Gerwig}.
\end{alltt}

\end{tcolorbox}

\begin{tcolorbox}[
  colback=gray!8, colframe=black!60,
  title={\textbf{Example 2} \small (2024/05/31)},
]
\small
\textbf{Question:} What is the name of the new restaurant that opened in Rehoboth Beach, Delaware?

\textbf{Evidence:} $\emptyset$ 

\textbf{OLMoE (no evidence)} \hfill \textsc{hallucinated}
\begin{alltt}\small
  \textcolor{red}{As of May 31, 2024, there isn't any 
  widely reported new restaurant 
  opening in Rehoboth Beach, Delaware, 
  at least not under the name you 
  provided. Rehoboth Beach is a 
  popular beach destination in Delaware, 
  and many restaurants come and go. 
  If you have more information about 
  the restaurant you're}
\end{alltt}

\textbf{OLMoE (with evidence)} \hfill \textsc{hallucinated}
\begin{alltt}\small
  \textcolor{red}{As of May 31, 2024, there is no new 
  restaurant that has officially 
  opened in Rehoboth Beach, Delaware, 
  according to available information. 
  However, there are several 
  restaurants in the area that have 
  been operating for years, such as 
  The Salty Dog, The Beachcomber, and 
  The Reh}
\end{alltt}

\textbf{Gemma 4 (no evidence)} \hfill \textsc{grounded}
\begin{alltt}\small
  I do not have information on a 
  specific new restaurant opening in 
  Rehoboth Beach, Delaware, as your 
  request does not specify a 
  particular name or recent opening 
  date. Please provide more details if 
  you are looking for a specific 
  establishment.
\end{alltt}

\textbf{Gemma 4 (with evidence)} \hfill \textsc{grounded}
\begin{alltt}\small
  Please provide the evidence 
  mentioned in your prompt, as it was 
  not included. Without the text, I 
  cannot identify the specific 
  restaurant you are referring to.
\end{alltt}

\end{tcolorbox}

\begin{tcolorbox}[
  colback=gray!8, colframe=black!60,
  title={\textbf{Example 3} \small (2024/01/05)},
]
\small
\textbf{Question:} What percentage of US adults have received the latest Covid-19 vaccine?

\textbf{Evidence:} Just 19\% of adults and 8\% of children have gotten the latest Covid-19 vaccine, CDC data shows. This comes as respiratory virus activity is high and rising across the US.

\textbf{OLMoE (no evidence)} \hfill \textsc{hallucinated}
\begin{alltt}\small
  \textcolor{red}{As of January 05, 2024}, \textcolor{red}{a significant 
  percentage} of US adults have received 
  the latest COVID-19 vaccine. 
  According to data from the Centers 
  for Disease Control and Prevention 
  (CDC), \textcolor{red}{as of January 2, 2024}, 
  \textcolor{red}{approximately 70.8\% of adults in the 
  United States have received at least 
  one dose of a COVID}
\end{alltt}

\textbf{OLMoE (with evidence)} \hfill \textsc{grounded}
\begin{alltt}\small
  19\% of US adults have received the 
  latest Covid-19 vaccine, according 
  to CDC data.
\end{alltt}

\textbf{Gemma 4 (no evidence)} \hfill \textsc{hallucinated}
\begin{alltt}\small
  \textcolor{red}{As of January 2024}, \textcolor{red}{there is no single 
  official percentage for the "latest" 
  COVID-19 vaccine across all US adults
  , as uptake varies by age group and 
  specific booster type.} However, CDC 
  data indicates that \textcolor{red}{roughly **22\%**} 
  of all US adults have received an 
  updated 202
\end{alltt}

\textbf{Gemma 4 (with evidence)} \hfill \textsc{grounded}
\begin{alltt}\small
  19\%
\end{alltt}

\end{tcolorbox}

\begin{tcolorbox}[
  colback=gray!8, colframe=black!60,
  title={\textbf{Example 4} \small (2024/01/05)},
]
\small
\textbf{Question:} Who beat 16-year-old darts prodigy Luke Littler to become the PDC world champion?

\textbf{Evidence:} British teen Littler, nicknamed "The Nuke", defeated former world No.1s Cross and van Gerwen to become the youngest-ever world championship finalist, before losing to Humphries. The 28-year-old new world champ took five consecutive sets to win 7-4 at London's Alexandra Palace on Wednesday.

\textbf{OLMoE (no evidence)} \hfill \textsc{hallucinated}
\begin{alltt}\small
  \textcolor{red}{As of January 05, 2024,} 16-year-old 
  darts prodigy Luke Littler did not 
  yet win the \textcolor{red}{PDC (Professional Darts 
  Corporation)} world championship. \textcolor{red}{The 
  PDC world champion for the 2023 
  season was Peter Wright, who 
  defeated Mark Webster in the final. 
  Peter Wright is a 38-}
\end{alltt}

\textbf{OLMoE (with evidence)} \hfill \textsc{hallucinated}
\begin{alltt}\small
  Luke Littler, also known as "The 
  Nuke," was defeated by another 
  player in the PDC world championship. 
  \textcolor{red}{Chris "Welly"} Humphries, a 
  \textcolor{red}{32-year-old} player, won the match 7-4 
  at London's Alexandra Palace on 
  Wednesday.
\end{alltt}

\textbf{Gemma 4 (no evidence)} \hfill \textsc{hallucinated}
\begin{alltt}\small
  \textcolor{red}{No one has beaten Luke Littler for 
  the PDC World Championship yet; the 
  2024 tournament is currently ongoing. 
  Luke Littler is a finalist/competitor 
  in the event, but the champion has 
  not been crowned.}
\end{alltt}

\textbf{Gemma 4 (with evidence)} \hfill \textsc{grounded}
\begin{alltt}\small
  Luke Humphries.
\end{alltt}

\end{tcolorbox}

\begin{tcolorbox}[
  colback=gray!8, colframe=black!60,
  title={\textbf{Example 5} \small (2025/04/04)},
]
\small
\textbf{Question:} More than 240 people fell ill recently due to a norovirus
  outbreak on which luxury cruise ship?

\textbf{Evidence:} The outbreak aboard the Queen Mary 2 comes on the heels of
  the worst year for cruise ship-based gastrointestinal outbreaks in over a
  decade.

\textbf{OLMoE (no evidence)} \hfill \textsc{hallucinated}
\begin{alltt}\small
  \textcolor{red}{As of April 03, 2025, there has been 
  no reported norovirus outbreak on a 
  luxury cruise ship involving more 
  than 240 people falling ill. The 
  norovirus is a highly contagious 
  virus that can cause gastroenteritis, 
  which includes symptoms such as 
  nausea, vomiting, diarrhea, and 
  stomach cramps. It is common on}
\end{alltt}

\textbf{OLMoE (with evidence)} \hfill \textsc{hallucinated}
\begin{alltt}\small
  The luxury cruise ship involved in 
  the norovirus outbreak that affected 
  more than 240 people was the Queen 
  Mary 2. \textcolor{red}{This incident occurred in 
  early 2023, as mentioned in your 
  evidence.}
\end{alltt}

\textbf{Gemma 4 (no evidence)} \hfill \textsc{hallucinated}
\begin{alltt}\small
  The norovirus outbreak occurred on 
  the \textcolor{red}{**Celebrity Eclipse**}.
\end{alltt}

\textbf{Gemma 4 (with evidence)} \hfill \textsc{grounded}
\begin{alltt}\small
  Queen Mary 2
\end{alltt}

\end{tcolorbox}

\subsection{LLM-as-a-Judge Label Validation (RQ4)}\label{subsec:llm-judge-validation}

The training supervision pipeline described in
Section~\ref{subsec:training-supervision} relies on LLM-as-a-judge labels as
ground truth. In this section, we validate the quality of these labels against
human annotations and compare them with a simpler weak labeling heuristic based
on lexical and semantic overlap metrics.

\paragraph{Weak labeling heuristic.}
Lexical and semantic overlap metrics (\eg BLEU, ROUGE, and BERTScore) are
commonly used to evaluate hallucination in LLM
outputs~\cite{ji2023survey, alansari2026large}. However, both these statistical
metrics often fail to assess the factuality and
faithfulness of generated text and lack robustness in aligning with human
judgments of hallucination~\cite{ji2023survey, alansari2026large}. We observed
this limitation empirically; the separability
between hallucinated and grounded answers remained low across all three
metrics. We nonetheless construct a weak labeling heuristic by computing
F1-optimal thresholds for each metric (estimated from base vs.\ RAG
separability on the training data) and averaging the resulting binary labels
into a weak hallucination label. The inadequacy of this heuristic motivates the
LLM-as-a-judge approach described in
Section~\ref{subsec:training-supervision}, which we validate below.

\paragraph{Human validation study.}
To assess label quality, we drew a stratified sample of 200 answers from the
RealtimeQA training data (January 2024--December 2025). The sampling design
consists of a $2{\times}2$ contingency stratification: for each host model
(OLMoE and Gemma), we sample 25 answers from each cell of the
weak-label $\times$ LLM-label contingency table (both agree-grounded,
both agree-hallucinated, weak-hallucinated/LLM-grounded, and
weak-grounded/LLM-hallucinated), yielding $25 \times 4 \times 2 = 200$
samples. A single human annotator evaluated each sample, labeling it as
hallucinated ($1$) or grounded ($0$) based on the question, the reference
evidence, and the generated answer. The annotator had no access to the labels
produced by either the weak heuristic or the LLM judge, ensuring that the
validation is not biased by either labeling approach.

\paragraph{Results.}
Table~\ref{tab:judge_validation_confusion} reports the confusion matrices
between human annotations and both labeling approaches, pooled across both host
models. The LLM judge achieves substantially higher agreement with human
annotations, with 27 misclassifications out of 200 (86.5\,\% accuracy),
compared to 81 misclassifications for the weak heuristic (59.5\,\% accuracy).
The LLM judge's errors are also more balanced: 20 false hallucinated labels
(grounded by human, but hallucinated by the LLM judge) and 7 false grounded labels,
whereas the weak heuristic exhibits 47 false hallucinated and 34 false grounded
labels. This confirms that the LLM-as-a-judge approach produces more reliable
training labels than the metric-based heuristic.

\begin{table}

\caption{Confusion matrix between human annotations and LLM-as-a-judge labels/Weak labeling on the 200-sample validation set (pooled across both host models, 25 per $2{\times}2$ contingency cell per model).}
\label{tab:judge_validation_confusion}
\centering
\begin{tabular}{lccc}
\toprule
  LLM-as-a-judge & Grounded & Hallucinated & Total \\
  Human &  &  \\
\midrule
  Grounded & 93 & 20 & 113 \\
  Hallucinated & 7 & 80 & 87 \\
  Total & 100 & 100 & 200 \\
\midrule
\midrule
  Weak heuristic & Grounded & Hallucinated & Total \\
Human &  &  \\
\midrule
  Grounded & 66 & 47 & 113\\
  Hallucinated & 34 & 53 & 87 \\
  Total & 100 & 100 & 200 \\
\bottomrule
\end{tabular}
\end{table}

\paragraph{Confidence distributions.}
Tables~\ref{tab:confidence-distribution-hallucinated}
and~\ref{tab:confidence-distribution-grounded} report the distribution of
LLM-judge labels across the weak heuristic's confidence scores (the calibrated
hallucination probability) for hallucinated and grounded labels, respectively.
For hallucinated labels, the judge's labels are heavily concentrated in the
$[0.8, 1.0]$ bin (74.4\,\% for OLMoE, 82.0\,\% for Gemma), indicating a good
level of agreement flagging hallucinations. For grounded labels, confidence is
concentrated in the $[0.0, 0.2[$ bin (24.9\,\% for OLMoE, 75.2\,\% for Gemma)
and the $[0.2, 0.4[$ bin (67.5\,\% for OLMoE, 12.8\,\% for Gemma), indicating
high confidence (low hallucination probability) for grounded answers. The clear
separation between the two distributions suggests the LLM judge is well
calibrated and that ambiguous cases (confidence near 0.5) are relatively rare.
However, high disagreement between the weak heuristic and the LLM judge is
observed in some cases, which highlights the cases where the weak heuristic
fails to label hallucinations while asserting high confidence.

\begin{table}[htbp]
\centering
\caption{Confidence distribution, based on the weak labeling approach, for
  hallucinated labels (label=1)}
  \label{tab:confidence-distribution-hallucinated}
  \begin{tabular}{lrr}
    \toprule
    Confidence & OLMoE (\%) & Gemma (\%) \\
    \midrule
    {[0.0, 0.2[} & 36 (0.9\,\%) & 220 (9.6\,\%) \\
    {[0.2, 0.4[} & 598 (15.5\,\%) & 88 (3.8\,\%) \\
    {[0.4, 0.6[} & 230 (6.0\,\%) & 33 (1.4\,\%) \\
    {[0.6, 0.8[} & 121 (3.1\,\%) & 71 (3.1\,\%) \\
    {[0.8, 1.0]} & 2862 (74.4\,\%) & 1877 (82.0\,\%) \\
    \midrule
    \textbf{Total} & \textbf{3847} (\textbf{100\,\%)} & \textbf{2289} (\textbf{100\,\%)} \\
    \bottomrule
  \end{tabular}
\end{table}

\begin{table}[htbp]
\centering
\caption{Confidence distribution, based on the weak labeling approach, for
  grounded labels (label=0)}
  \label{tab:confidence-distribution-grounded}
  \begin{tabular}{lrr}
    \toprule
    Confidence & OLMoE (\%) & Gemma (\%) \\
    \midrule
    {[0.0, 0.2[} & 196 (24.9\,\%) & 1764 (75.2\,\%) \\
    {[0.2, 0.4[} & 531 (67.5\,\%) & 301 (12.8\,\%) \\
    {[0.4, 0.6[} & 44 (5.6\,\%) & 50 (2.1\,\%) \\
    {[0.6, 0.8[} & 6 (0.8\,\%) & 42 (1.8\,\%) \\
    {[0.8, 1.0]} & 10 (1.3\,\%) & 189 (8.1\,\%) \\
    \midrule
    \textbf{Total} & \textbf{787} (\textbf{100\,\%)} & \textbf{2346} (\textbf{100\,\%)} \\
    \bottomrule
  \end{tabular}
\end{table}

\paragraph{Label distribution.}
Table~\ref{tab:label-distribution} reports the label distribution by host model
and evidence condition. OLMoE exhibits a strong class imbalance, with
83.0\,\% of answers labeled as hallucinated overall (99.7\,\% for base answers
without evidence, 66.2\,\% for answers generated with evidence). Gemma is
more balanced, with 49.4\,\% hallucinated overall (85.4\,\% base, 13.3\,\%
with evidence). The contrast between base and evidence conditions creates a
natural label diversity that benefits classifier training, though the class
imbalance for OLMoE should be considered when interpreting threshold-dependent
metrics such as F1 (cf.\ Section~\ref{sec:experiments}).


\begin{table}[htbp]
\centering
\caption{Label distribution by model and evidence condition}
\label{tab:label-distribution}
\begin{tabular}{lcccc}
\toprule
Condition & N & Halluc. & Grounded & Halluc.\,\% \\
\midrule
  \multicolumn{5}{c}{OLMoE-1B-7B} \\
\midrule
Base & 2318 & 2312 & 6 & 99.7\,\% \\
Evidence & 2318 & 1535 & 783 & 66.2\,\% \\
\textbf{Overall} & \textbf{4636} & \textbf{3847} & \textbf{789} & \textbf{83.0\,\%} \\
\midrule
  \multicolumn{5}{c}{Gemma-4-26B} \\
\midrule
Base & 2318 & 1980 & 338 & 85.4\,\% \\
Evidence & 2318 & 309 & 2009 & 13.3\,\% \\
\textbf{Overall} & \textbf{4636} & \textbf{2289} & \textbf{2347} & \textbf{49.4\,\%} \\
\bottomrule
\end{tabular}
\end{table}

\subsection{Per-token hallucination detection (RQ1 - Cont.)}\label{sec:per-token-cont}

Table~\ref{tab:token_level_results_F1} reports token-level F1. The
threshold-dependent results reveal greater variability across classifier
families than the AUROC analysis (Section~\ref{subsec:per-token}). IE (MLP) and
IE (LR) remain the most consistent performers, with IE (MLP) achieving the
highest average F1 on OLMoE and IE (LR) on Gemma. In contrast, IE (XGBoost),
the strongest variant under AUROC, suffers a sharp F1 degradation on Gemma,
indicating that its F1-optimized threshold does not generalize across datasets.
LC (att.) produces zero F1 on RealTimeQA for both models, reflecting threshold
instability rather than a lack of discriminative signal.

\begin{table}[htb]
\caption{Token-level results (F1) across datasets and host models.}
\label{tab:token_level_results_F1}
\centering
\footnotesize
\setlength{\tabcolsep}{3pt}
\begin{tabular}{@{}lccccccc@{}}
\toprule
  & FQA & NQO & RTQA & SQuAD & TQA & Rank & Avg \\
\midrule
  \multicolumn{8}{c}{Gemma-4-26B} \\
\midrule
IE (LR) & 0.648 & 0.629 & 0.751 & 0.705 & 0.730 & \textbf{2.6} & \textbf{0.693} \\
IE (MLP) & 0.646 & 0.625 & 0.681 & \textbf{0.718} & 0.706 & 4.0 & 0.675 \\
IE (RF) & 0.571 & 0.575 & 0.756 & 0.706 & \textbf{0.774} & 3.2 & 0.676 \\
IE (Transf.) & 0.586 & 0.544 & 0.726 & 0.653 & 0.671 & 5.2 & 0.636 \\
IE (XGB) & 0.175 & 0.116 & \textbf{0.782} & 0.139 & 0.168 & 6.6 & 0.276 \\
LC (att.) & 0.649 & \textbf{0.645} & 0.000 & 0.602 & 0.720 & 4.2 & 0.523 \\
LC (hid.) & \textbf{0.652} & 0.627 & 0.738 & 0.643 & 0.727 & 3.2 & 0.678 \\
Entropy & 0.332 & 0.381 & 0.307 & 0.426 & 0.462 & 7.0 & 0.382 \\

 \midrule
  \multicolumn{8}{c}{OLMoE-1B-7B} \\
\midrule
IE (LR) & 0.733 & \textbf{0.781} & 0.784 & \textbf{0.714} & 0.760 & 2.6 & 0.754 \\
IE (MLP) & \textbf{0.783} & 0.779 & \textbf{0.794} & 0.688 & 0.825 & \textbf{1.6} & \textbf{0.774} \\
IE (RF) & 0.732 & 0.604 & 0.631 & 0.508 & \textbf{0.826} & 5.0 & 0.660 \\
IE (Transf.) & 0.741 & 0.595 & 0.657 & 0.542 & 0.819 & 4.8 & 0.671 \\
IE (XGB) & 0.629 & 0.565 & 0.753 & 0.555 & 0.564 & 6.6 & 0.613 \\
LC (att.) & 0.763 & 0.773 & 0.000 & 0.616 & 0.815 & 4.2 & 0.594 \\
LC (hid.) & 0.684 & 0.658 & 0.144 & 0.628 & 0.756 & 5.4 & 0.574 \\
Entropy & 0.645 & 0.672 & 0.603 & 0.595 & 0.721 & 5.8 & 0.647 \\
\bottomrule
\end{tabular}
\end{table}

\subsection{Comparison to baselines (RQ2 - Cont.)}\label{sec:comparison-baselines-cont}

Table~\ref{tab:answer_level_results_f1} reports answer-level F1. IE (MLP)
achieves the highest average F1 on both host models, consistent with its strong
AUROC performance. IE (XGBoost) again degrades sharply on Gemma, confirming the
threshold sensitivity observed at the token level. On OLMoE, F1 values are
inflated across most methods (many exceeding 0.94), likely due to the extreme
label imbalance on this model (Table~\ref{tab:label-distribution});
consequently, the F1 ranking is less informative than AUROC. Sampling-based
methods exhibit low F1 on Gemma, with SelfCheckGPT (Prompt) reaching only
0.129, but recover on OLMoE. The full inference benchmark table is provided in
Table~\ref{tab:inference_benchmark}; see
Section~\ref{subsec:comparison-baselines} for discussion.

\begin{table}[htb]
\caption{Answer-level results (F1) across datasets and host models.}
\label{tab:answer_level_results_f1}
\footnotesize
\setlength{\tabcolsep}{2pt}
\begin{tabular}{@{}lccccccc@{}}
\toprule
  & FQA & NQO & RTQA & SQuAD & TQA & Rank & Avg \\
\midrule
  \multicolumn{8}{c}{Gemma-4-26B} \\
\midrule
HaluNet & 0.778 & 0.769 & 0.849 & 0.825 & 0.897 & 3.2 & 0.824 \\
IE (LR) & 0.790 & 0.763 & 0.849 & 0.873 & 0.904 & 2.4 & 0.836 \\
IE (MLP) & \textbf{0.828} & 0.780 & \textbf{0.876} & \textbf{0.890} & 0.868 & \textbf{2.0} & \textbf{0.849} \\
IE (RF) & 0.689 & 0.679 & 0.809 & 0.837 & \textbf{0.912} & 4.6 & 0.785 \\
IE (Transf.) & 0.751 & \textbf{0.787} & 0.840 & 0.760 & 0.884 & 3.6 & 0.804 \\
IE (XGB) & 0.444 & 0.410 & 0.804 & 0.472 & 0.636 & 9.6 & 0.553 \\
LC (att.) & 0.722 & 0.736 & 0.705 & 0.601 & 0.654 & 7.0 & 0.684 \\
LC (hid.) & 0.722 & 0.736 & 0.705 & 0.710 & 0.654 & 6.6 & 0.705 \\
Entropy & 0.585 & 0.658 & 0.722 & 0.689 & 0.735 & 7.8 & 0.678 \\
Perplexity & 0.317 & 0.395 & 0.332 & 0.424 & 0.561 & 12.4 & 0.405 \\
SCGPT (NLI) & 0.354 & 0.383 & 0.401 & 0.306 & 0.183 & 12.6 & 0.325 \\
SCGPT (Prompt) & 0.221 & 0.174 & 0.133 & 0.077 & 0.037 & 14.0 & 0.129 \\
SemEnergy & 0.643 & 0.675 & 0.633 & 0.588 & 0.587 & 9.2 & 0.625 \\
SemUncert & 0.594 & 0.590 & 0.604 & 0.597 & 0.581 & 10.0 & 0.593 \\

 \midrule
  \multicolumn{8}{c}{OLMoE-1B-7B} \\
\midrule
HaluNet & 0.963 & 0.975 & 0.947 & 0.884 & 0.966 & 5.2 & 0.947 \\
IE (LR) & 0.955 & 0.973 & 0.931 & 0.830 & 0.963 & 9.0 & 0.930 \\
IE (MLP) & \textbf{0.973} & \textbf{0.981} & \textbf{0.950} & 0.875 & \textbf{0.978} & \textbf{1.8} & \textbf{0.951} \\
IE (RF) & 0.969 & 0.952 & 0.847 & 0.864 & 0.977 & 8.1 & 0.922 \\
IE (Transf.) & 0.972 & 0.972 & 0.905 & 0.871 & 0.976 & 6.6 & 0.939 \\
IE (XGB) & 0.952 & 0.957 & 0.877 & 0.857 & 0.919 & 11.0 & 0.913 \\
LC (att.) & 0.972 & 0.976 & 0.930 & 0.872 & 0.976 & 4.7 & 0.945 \\
LC (hid.) & \textbf{0.973} & 0.687 & 0.929 & 0.780 & 0.976 & 7.7 & 0.869 \\
Entropy & 0.969 & 0.972 & 0.935 & 0.876 & 0.976 & 5.2 & 0.946 \\
Perplexity & 0.968 & 0.972 & 0.934 & 0.874 & 0.974 & 6.9 & 0.944 \\
SCGPT (NLI) & 0.738 & 0.674 & 0.716 & 0.449 & 0.318 & 13.0 & 0.579 \\
SCGPT (Prompt) & 0.366 & 0.290 & 0.326 & 0.154 & 0.066 & 14.0 & 0.241 \\
SemEnergy & 0.961 & 0.974 & 0.939 & \textbf{0.896} & 0.975 & 5.2 & 0.949 \\
SemUncert & 0.970 & 0.974 & 0.930 & 0.870 & 0.974 & 6.6 & 0.944 \\
\bottomrule
\end{tabular}
\end{table}

\begin{table}
  \caption{
    Inference time (per 100 tokens) and peak GPU memory for each detection
    method, measured on 50 RealTimeQA questions per host model. InnerExpert
    times include MoE feature extraction (SVD-based hidden scores, attention
    scores, expert routing signals) as well as classifier inference.
  }
  \label{tab:inference_benchmark}
  \centering
  \footnotesize
  \setlength{\tabcolsep}{2pt}
  \begin{tabular}{@{}lcccc@{}}
    \toprule
    & \multicolumn{2}{c}{OLMoE-1B-7B} & \multicolumn{2}{c}{Gemma-4-26B} \\
    Method & \makecell{Time per \\ 100 tok (s)} & \makecell{Peak GPU \\ (GB)} & \makecell{Time per \\ 100 tok (s)} & \makecell{Peak GPU \\ (GB)} \\
    \midrule
    Vanilla & 1.146 & 12.660 & 3.825 & 46.810 \\
    Logit Entropy & 1.152 & 12.670 & 3.830 & 46.880 \\
    Perplexity & 1.152 & 12.670 & 3.830 & 47.010 \\
    LC (att.) & 1.198 & 12.680 & 3.965 & 46.880 \\
    LC (hid.) & 1.246 & 12.670 & 4.133 & 46.830 \\
    HaluNet & 1.257 & 12.680 & 4.128 & 47.050 \\
    Router Entropy & 2.829 & 12.820 & 7.677 & 47.470 \\
    Exp. Hidden & 2.829 & 12.820 & 7.677 & 47.470 \\
    Exp. Similarity & 2.829 & 12.820 & 7.677 & 47.470 \\
    Exp. Entropy & 2.829 & 12.820 & 7.677 & 47.470 \\
    Usage Gini & 2.829 & 12.820 & 7.677 & 47.470 \\
    Inv. Herfindahl & 2.829 & 12.820 & 7.677 & 47.470 \\
    IE (LR) & 3.493 & 13.080 & 9.376 & 48.110 \\
    IE (MLP) & 3.492 & 13.080 & 9.377 & 48.110 \\
    IE (RF) & 3.537 & 13.080 & 9.409 & 48.110 \\
    IE (XGBoost) & 3.498 & 13.080 & 9.384 & 48.110 \\
    IE (Transf.) & 3.495 & 13.110 & 9.384 & 48.180 \\
    SemUncert & 6.722 & 14.230 & 21.860 & 48.360 \\
    SemEnergy & 7.101 & 14.230 & 23.883 & 48.360 \\
    SCGPT (NLI) & 5.966 & 14.200 & 19.331 & 48.340 \\
    SCGPT (Prompt) & 6.108 & 26.850 & 19.562 & 59.340 \\
    \bottomrule
  \end{tabular}
\end{table}

\subsection{Signal contribution analysis (RQ3 - Cont.)}\label{sec:signal-analysis-cont}

Table~\ref{tab:signal_contribution_F1} reports F1 for individual signals
and Machine Learning classifiers. The combination benefit observed under AUROC
is preserved for IE (MLP), which achieves the highest F1 on both models at
both levels. However, IE (XGBoost) underperforms several individual signals at
F1 on both models, the opposite of the AUROC finding, again reflecting
threshold sensitivity rather than reduced discriminative power. 

\begin{table}[htb]
\caption{Average F1 for individual MoE signals, baselines, 
  \method (XGB). Each value is averaged across all five evaluation datasets
  per host model. Answer-level scores use the best aggregation per method. Methods
  without a token-level counterpart are marked ``---''.}
\label{tab:signal_contribution_F1}
\centering
\footnotesize
\setlength{\tabcolsep}{3pt}
\begin{tabular}{lcccc}
\toprule
 & \multicolumn{2}{r}{OLMoE-1B-7B} & \multicolumn{2}{r}{Gemma-4-26B} \\
  Method & Answer & Token & Answer & Token \\
\midrule
  Inv. Herfindahl       & 0.940 & \textbf{0.731} & 0.697 & 0.619 \\
  Exp. Entropy          & 0.881 & 0.639 & \textbf{0.748} & \textbf{0.671} \\
  Exp. Gini             & 0.869 & 0.597 & 0.745 & \textbf{0.671} \\
  Expert Hidden         & \textbf{0.945} & 0.698 & 0.707 & 0.667 \\
  Expert Similarity     & \textbf{0.945} & 0.634 & 0.740 & 0.523 \\
  Router Entropy        & 0.857 & \textbf{0.731} & 0.571 & 0.667 \\
\midrule
  Hid. Score            & 0.869 & 0.574 & \textbf{0.705} & \textbf{0.678} \\
  Att. Score            & 0.945 & 0.594 & 0.684 & 0.523 \\
  Logit Entropy         & \textbf{0.946} & \textbf{0.647} & 0.680 & 0.382 \\
  Perplexity            & 0.944 & --- & 0.405 & --- \\
\midrule
  IE (XGB)              & 0.913 & 0.613 & 0.553 & 0.276 \\
  IE (MLP)              & \textbf{0.951} & \textbf{0.774} & \textbf{0.849} & \textbf{0.675} \\
  HaluNet               & 0.947 & --- & 0.824 & --- \\
  SCGPT (NLI)           & 0.579 & --- & 0.325 & --- \\
  SCGPT (Prompt)        & 0.241 & --- & 0.129 & --- \\
  SemUncertainty        & 0.944 & --- & 0.593 & --- \\
  SemEnergy             & 0.949 & --- & 0.625 & --- \\
\bottomrule
\end{tabular}
\end{table}

\section{Future Work}\label{sec:future-work}

This work opens several avenues for future research. First, taking
inspiration from methods such as~\citet{emde2025shh}, one could certify the
$\tau$-lucidity of a model (\ie the guarantee that the model will not
hallucinate with probability $1-\tau$ within a given domain), building on
the threshold discussion in Section~\ref{subsec:architecture}. Second,
adapting approaches such as~\citet{fonseca2025safenudge} for hallucination
detection, or extending \method as a Controlled Text Generation method, is a
promising direction. Third, while the unsupervised ground truth annotation
approach is effective, it remains prone to noisy labels; more robust labeling
methods could be explored. Fourth, while our results support the use of
lightweight classifiers, the threshold sensitivity of XGBoost
(Section~\ref{sec:per-token-cont}) suggests that ad-hoc classifier
architectures or threshold calibration techniques could improve robustness.
Fifth, the unsupervised training pipeline naturally extends to online model
monitoring, enabling detection of hallucination rate drift over time.
Finally, transferring \method across domains, tasks, languages, and MoE
architectures without retraining is an important direction for scalability.


\end{document}